\documentclass[conference]{IEEEtran}
\IEEEoverridecommandlockouts
\usepackage{cite}
\usepackage{amsmath,amssymb,amsfonts}
\usepackage{algorithmic}
\usepackage{graphicx}
\usepackage{textcomp}
\usepackage{xcolor} 
\usepackage{booktabs}
\usepackage{url}
\usepackage{multirow} 
\usepackage{pifont}
\newcommand{\xmark}{\ding{55}}
\usepackage{microtype}
\usepackage{subcaption}
\usepackage{graphicx}

\usepackage{amsmath}
\usepackage{amssymb}
\usepackage{mathtools}
\usepackage{amsthm}
\def\BibTeX{{\rm B\kern-.05em{\sc i\kern-.025em b}\kern-.08em
    T\kern-.1667em\lower.7ex\hbox{E}\kern-.125emX}}

\newcommand{\Lvec}{\mathbf{L}}
\newcommand{\Lx}{L_x}
\newcommand{\Ly}{L_y}
\newcommand{\Lz}{L_z}

\begin{document}

\title{Physics-informed Attention-enhanced Fourier Neural Operator for Solar Magnetic Field Extrapolations}



\author{
\IEEEauthorblockN{Jinghao Cao}
\IEEEauthorblockA{\textit{Dept. Mechanical and} \\
\textit{Industrial Engineering} \\
\textit{NJIT}, Newark, USA\\
jc2687@njit.edu}
\and
\IEEEauthorblockN{Qin Li}
\IEEEauthorblockA{\textit{Dept. Physics} \\
\textit{NJIT}, Newark, USA\\
ql47@njit.edu}
\and
\IEEEauthorblockN{Mengnan Du}
\IEEEauthorblockA{\textit{Dept. Data Science} \\
\textit{NJIT}, Newark, USA\\
mengnan.du@njit.edu} \\
\and
\IEEEauthorblockN{Haimin Wang}
\IEEEauthorblockA{\textit{Dept. Physics} \\
\textit{NJIT}, Newark, USA\\
haimin.wang@njit.edu}
\and
\IEEEauthorblockN{Bo Shen*\thanks{*Dr. Bo Shen is the corresponding author.}}
\IEEEauthorblockA{\textit{Dept. Mechanical and} \\
\textit{Industrial Engineering} \\
\textit{NJIT}, Newark, USA\\
bo.shen@njit.edu}
}



\maketitle

\begin{abstract}
We propose \textbf{P}hysics-\textbf{i}nformed~\textbf{A}ttention-enhanced~Fourier~\textbf{N}eural~\textbf{O}perator~\textbf{(PIANO)}  to solve the Nonlinear Force-Free Field (NLFFF) problem in solar physics. Unlike conventional approaches that rely on iterative numerical methods, our proposed PIANO directly learns the 3D magnetic field structure from 2D boundary conditions. Specifically, PIANO integrates Efficient Channel Attention (ECA) mechanisms with Dilated Convolutions (DC), which enhances the model's ability to capture multimodal input by prioritizing critical channels relevant to the magnetic field's variations. Furthermore, we apply physics-informed loss by enforcing the force-free and divergence-free conditions in the training process so that our prediction is consistent with underlying physics with high accuracy. Experimental results on the ISEE NLFFF dataset show that our PIANO not only outperforms state-of-the-art neural operators in terms of accuracy but also shows strong consistency with the physical characteristics of NLFFF data across magnetic fields reconstructed from various solar active regions. The GitHub of this project is available \url{https://github.com/Autumnstar-cjh/PIANO}
\end{abstract}


\section{Introduction}
Solar magnetic field extrapolations have been pivotal in understanding the Sun’s activity and its influence on space weather, such as solar flares, coronal mass ejections, and geomagnetic storms. One of the most commonly employed techniques in solar magnetic field extrapolations is the Nonlinear Force-Free Field (NLFFF) approximation \cite{low1990, wiegelmann2008}, which assumes that the coronal magnetic field is in a quasi-static state where the Lorentz force vanishes. This approach is quite effective for modeling the solar corona and has been widely used to infer the three-dimensional structure of the corona. However, NLFFF methods face a critical challenge: computational inefficiency \cite{derosa2015influence}. Solving the associated non-linear equations involves iterative processes that require substantial computational resources, especially when applied to domains of high spatial resolution. This computational burden limits the applicability of NLFFF in scenarios requiring rapid or real-time analysis, such as operational space weather forecasting or large-scale parametric studies.

In recent years, artificial intelligence (AI) / machine learning (ML) methods have offered new capabilities for solar physics research \cite{AsensioRamos2023,shen2024deep}, including applications with NLFFF.  \cite{chifu2021} presented a method to retrieve 3D loops from NLFFF based on the Convolutional neural network (CNN).  \cite{zhang2024Machine} proposed a grid-free machine-learning algorithm and demonstrated the efficient capability of deriving numerical solutions for NLFFF. However, none of the above methods considers integrating physics laws into neural networks. As an alternative, Physics-informed neural networks (PINNs) can include the governing physics described by general nonlinear partial differential equations (PDEs) into neural networks \cite{raissi2019physics,karniadakis2021physics}. \cite{Zhang2024} proposed a PINN extrapolation method that leverages machine learning techniques to reconstruct coronal magnetic fields.  \cite{jarolim2023} used PINN to integrate observational data and the NLFFF model to predict NLFFF extrapolations, and their following work \cite{jarolim2024} utilized multi-height magnetic field measurements to advance solar magnetic field extrapolations.

Despite considerable progress beyond conventional physics-based modeling, PINNs have a notable limitation: they are designed to address a particular PDE or a specific instance of a problem (i.e., with fixed boundary/initial condition and coefficients in the PDE).  In contrast, neural operators \cite{li2021, Nikola2023, azizzadenesheli2024neural} offer a more generalized framework by directly learning a mapping from function spaces of inputs to outputs, effectively acting as a solver for any instance of the PDE, not just a pre-specified one. Therefore, neural operators are the top options for scientific and engineering problems. While in the domain of coronal magnetic field extrapolations using NLFFF, there is no documented application of Fourier neural operators (FNOs) to derive the solution of NLFFF. Given the significant computational and time demands of NLFFF and its pressing need for real-time application, there is a need to utilize FNOs to accelerate NLFFF. 

However, there are two challenges to applying existing FNO-based methods to NLFFF directly (1) \underline{\textit{multimodal input}}: in NLFFF, we have different types of inputs, such as boundary conditions (in the format of images) and physical scalars (in the format of a vector), etc; (2) \underline{\textit{physics laws}}: the nonlinear PDEs in NLFFF need to be included in FNO to align with real-world physics.

To address these two challenges, we propose \textbf{P}hysics-\textbf{i}nformed~\textbf{A}ttention-enhanced~Fourier~\textbf{N}eural~\textbf{O}perator~\textbf{(PIANO)}. Specifically, our PIANO can take both images and scalars as input to the model.  The scalar vector is processed through an Efficient Channel Attention (ECA) block \cite{wang2020} with Dilated Convolution \cite{yu2016}, which dynamically identifies and enhances the scalar-specific features most relevant to the magnetic field extrapolation. After this attention-based refinement, the scalar features are element-wise added to the input boundary conditions, ensuring that the model effectively integrates scalar-specific information with the observed boundary data. We also enforced NLFFF conditions in our loss function to maintain physics consistency. Furthermore, we also develop a two-phase training procedure to improve the overall performance.

In our experiments, we include state-of-the-art (SOTA) neural operator methods such as FNO \cite{li2021}, GLFNO \cite{du2024global}, UFNO \cite{wen2022}, GeoFNO \cite{li2024}, General Neural Operator Transformer (GNOT) \cite{hao2023}, FNOMIO \cite{JIANG2024110392} and PINO \cite{Jeon2025} as baseline methods for comparison.  In summary, our \textbf{contributions} are as follows:
\begin{itemize}
    \item We propose a novel neural operator called PIANO to obtain solutions from NLFFF, featuring multimodal input and incorporating physics laws.

    \item We evaluate our PIANO on the ISEE NLFFF database \cite{kusano2021} to demonstrate the effectiveness and efficiency of PIANO.  Our PIANO achieves the best test accuracy compared to SOTA baselines.
    
    \item In addition to the evaluation metrics from an AI/ML perspective, evaluation from a physics perspective is conducted to verify that our prediction from PIANO is reliable.
\end{itemize}

\section{Related Work}

\textbf{Solar Magnetic Field Modeling.} Accurately modeling solar magnetic fields is essential for uncovering the mechanisms driving solar activity and predicting its influence on space weather and Earth's environment. Traditional methods primarily rely on solving Magneto-HydroDynamic Equations (MHD) \cite{zhdanov2002plasma, Yamasaki2022, Inoue2023}, offering a comprehensive approach by capturing the interplay between magnetic fields and plasma dynamics, enabling large-scale simulations of phenomena like solar flares and coronal mass ejections. NLFFF, on the other hand, focuses on reconstructing the magnetic field in regions such as the solar corona, where magnetic forces dominate over plasma forces  \cite{wiegelmann2008, schrijver2006nlfff}. This makes NLFFF particularly effective for providing detailed and accurate insights into the intricate magnetic structures driving solar activity. However, as mentioned before, these methods are computationally expensive and sensitive to data quality and boundary conditions.


\textbf{Neural Operators.} FNO \cite{li2021} is a type of neural operator specifically designed to tackle parameterized PDEs efficiently. By leveraging the Fourier transform, FNO operates on functions by learning a mapping in the Fourier space, which allows for capturing the global interactions of the input functions. Several variations of the FNOs have been developed to enhance their capability across different scientific and engineering domains. GLFNO \cite{du2024global} improves the expressiveness of FNO by incorporating global and local branches, enabling global and local features learning simultaneously and enhancing its ability to capture fine details. UFNO \cite{wen2022} extends the FNO by introducing a unified spectral representation that leverages both Fourier and wavelet transforms. This hybrid approach enables UFNO to capture both global frequency information and localized spatial features. GeoFNO \cite{li2024} extends FNO to geophysical applications by effectively capturing the solution structures of various partial differential equations (PDEs), making it a powerful tool for modeling physical phenomena across different scales. GNOT \cite{hao2023} integrates transformer-based attention mechanisms into the neural operator framework, enabling it to learn mappings across multiple challenging datasets from diverse domains. FNOMIO \cite{JIANG2024110392} introduces a hybrid approach that combines multiple integral operators with Fourier transformations, enhancing the accuracy and stability of predictions in complex porous media flow simulations. PINO \cite{Jeon2025} uses PINO, which is a U-shaped neural operator architecture and incorporates physics information, to learn the solution operator that maps 2D photospheric vector magnetic fields to 3D nonlinear force-free fields.  

\begin{figure}[!htbp]
    \centering
    \includegraphics[width=0.5\textwidth]{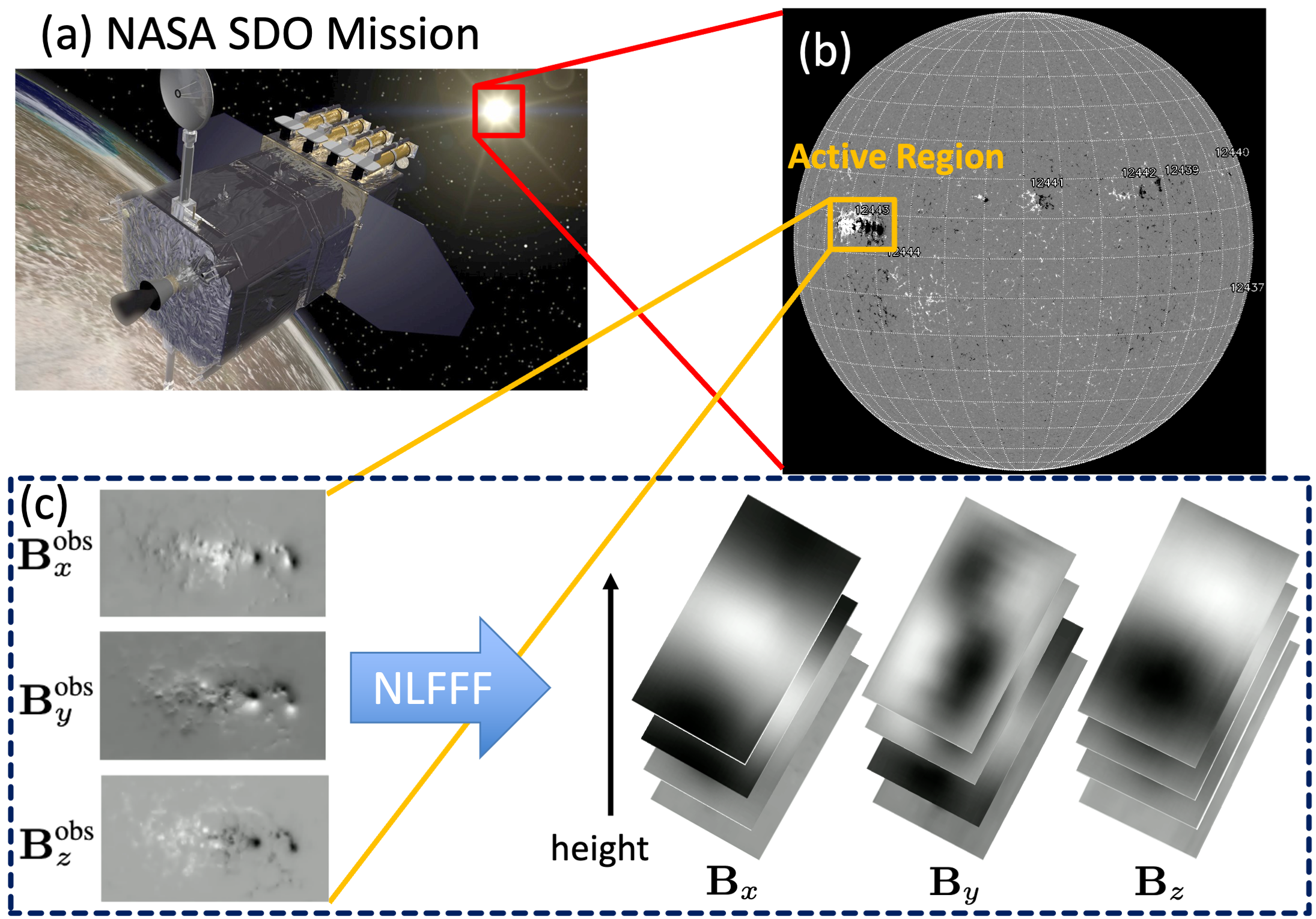}
    \caption{Overall pipeline of NLFFF. (a) NASA SDO mission; (b) Observed photospheric magnetogram; (c) the boundary conditions $\mathbf{B}^{\text{obs}}$ from an active region in magnetogram for NLFFF to obtain $\mathbf{B}$ extrapolated along the height.}
    \vspace{-0.4cm}
    \label{fig:Pipeline}
\end{figure}

\section{Preliminary}

\subsection{NLFFF} 
We show the overall pipeline of NLFFF in Fig.~\ref{fig:Pipeline}. The NLFFF model describes the magnetic field $\mathbf{B}=\{\mathbf{B}_x, \mathbf{B}_y, \mathbf{B}_z\}$ in the solar corona under the assumptions that the field is force-free and divergence-free, where $\mathbf{B}_x, \mathbf{B}_y, \mathbf{B}_z$ are 3D cubes. These conditions are expressed as follows:  
  
  \textbf{Force-Free Condition:}
    \begin{equation}
    (\nabla \times \mathbf{B}) \times \mathbf{B} = 0,
    \label{eq:force_free}
    \end{equation}
    which implies that the Lorentz force is zero. Physically, this means that the magnetic field is sufficiently dominant to neglect other forces such as pressure gradients and gravity.

   \textbf{Divergence-Free Condition:}
    \begin{equation}
    \nabla \cdot \mathbf{B} = 0,
    \label{eq:div_free}
    \end{equation}
    which ensures the solenoidal property of the magnetic field, as required by Maxwell's equations.

   \textbf{Boundary Conditions:}
    To solve the NLFFF equations, boundary conditions are required on the photospheric surface:
    \begin{equation}
    \mathbf{B} \big|_{\partial \Omega} = \mathbf{B}^{\text{obs}},
    \label{eq:boundary_conditions}
    \end{equation}
    where \(\mathbf{B}\) represents the magnetic field, \(\partial \Omega\) refers to the boundary of the computational domain \(\Omega\). $\mathbf{B}^{\text{obs}}=\{\mathbf{B}^{\text{obs}}_x,\mathbf{B}^{\text{obs}}_y, \mathbf{B}^{\text{obs}}_z\}$ is the observed photospheric magnetogram from NASA's SDO mission \cite{pesnell2012solar}, where $\mathbf{B}^{\text{obs}}_x,\mathbf{B}^{\text{obs}}_y, \mathbf{B}^{\text{obs}}_z$ are 2D vector fields.

These equations~\eqref{eq:force_free},~\eqref{eq:div_free},~\eqref{eq:boundary_conditions}, describe a nonlinear and nonlocal system, requiring iterative numerical methods to extrapolate $\mathbf{B}$  along the height dimension. Traditional methods for reconstructing NLFFF solutions rely on iterative numerical schemes such as \cite{Gilchrist2013} or magnetofrictional relaxation \cite{inoue2013}. They are computationally expensive and sensitive to noise, making them less practical for high-resolution or real-time scenarios.

Our goal is to learn a neural operator that maps the observed photospheric boundary measurements $\mathbf{B}^{\text{obs}}$ (in our experiment, in the dimension $\mathbb{R}^{3\times 257 \times 513}$) to the extrapolated $\mathbf{B}$ (in our experiment, in the dimension $\mathbb{R}^{3\times 128 \times 257 \times 513}$, where $128$ is the height dimension). Formally, we seek a parameterized operator $\mathcal{G}_\theta$ such that
\begin{equation} \label{eq: task}
\mathcal{G}_\theta : \mathbf{B}^{\text{obs}} \; \mapsto \; \mathbf{B},
\end{equation}
where $\theta$ denotes the learnable weights in our model. 
\begin{figure*}[!htbp]
    \centering
    \includegraphics[width=\textwidth]{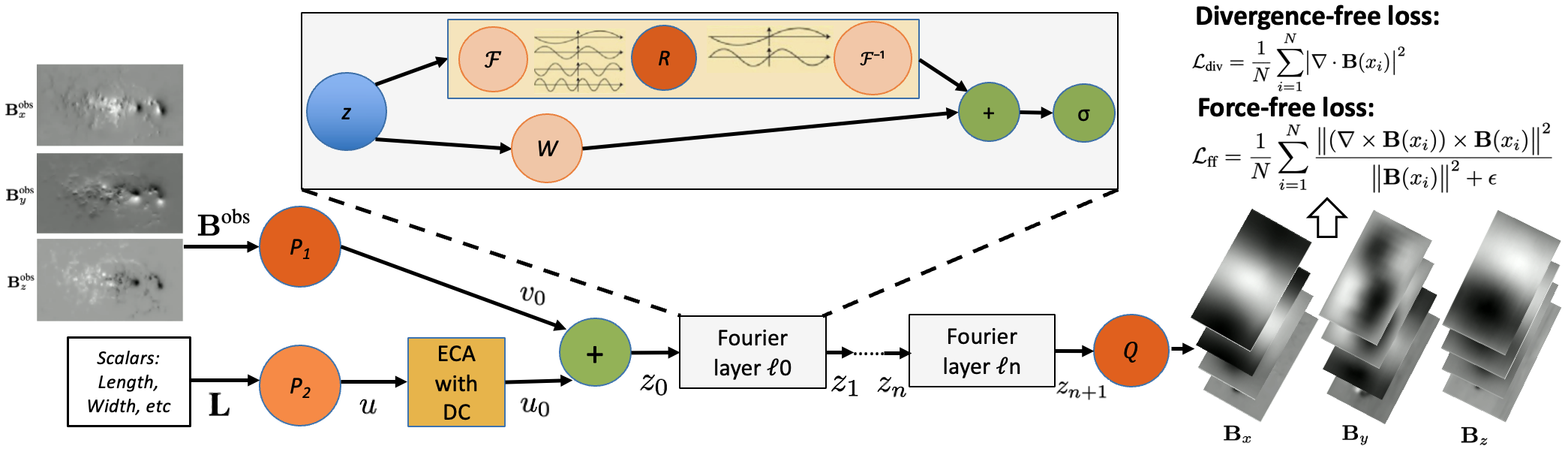}  
    \caption{\textbf{Architecture of PIANO.} The input consists of observed magnetic field $\mathbf{B}^{\text{obs}}$ and scalar parameters. $\mathbf{B}^{\text{obs}}$ is processed through the lifting layer $P_1$ to get feature $v_0$. Scalar parameters are processed by the lifting layer $P_2$ and an ECA with DC block to get the feature $u_0$. The aggregated features $z_0$ are passed through a series of Fourier layers. The projection layer $Q$ maps the $z_{n+1}$ to magnetic fields $\mathbf{B}_x, \mathbf{B}_y, \mathbf{B}_z$. The model includes  \(\mathcal{L}_{\text{div}}\) and \(\mathcal{L}_{\text{ff}}\), ensuring physics consistency in the magnetic fields.} \vspace{-0.5cm}
    \label{fig:PIANO}
\end{figure*}

\section{Method}
We propose a novel \textbf{P}hysics-\textbf{i}nformed~\textbf{A}ttention-enhanced~Fourier~\textbf{N}eural~\textbf{O}perator~\textbf{(PIANO)}  that leverages Fourier layer with multimodal input, as well as physics laws. The overall goal of our proposed PIANO is to reduce computational overhead for NLFFF without sacrificing accuracy. A schematic diagram of PIANO is shown in Fig.~\ref{fig:PIANO}. 

\subsection{Fourier Layer}

The Fourier integral operator \(\mathcal{K}\) is defined as:
\begin{equation*}
\mathcal{K}(\phi) z_i = \mathcal{F}^{-1} \left( R_\phi \cdot (\mathcal{F} z_i) \right),
\label{eq:layer1}
\end{equation*}
where \( \mathcal{F} \) and \( \mathcal{F}^{-1} \) denote the Fourier transform and its inverse, respectively.
 \( R_\phi \) is the Fourier transform of a periodic function \( \kappa : \bar{D} \to \mathbb{R}^{d_v \times d_v} \), parameterized by \(\phi \in \Theta_\mathcal{K}\). $z_i$ is the input function at layer \( i \in \{0,1,\dots,n\} \).

At each layer \( i \), the Fourier layer applies the Fourier integral operator \(\mathcal{K}\) and a pointwise nonlinearity:
   \begin{equation*}
z_{i+1} = \sigma\left((\mathcal{K}(\phi_i)z_i) + W_i(z_i)\right),
\label{eq:layer2}
\end{equation*}
    where \( W_i\) is a pointwise linear transformation and \(\sigma\) is a nonlinear activation function.

\subsection{Model Architecture}

\textbf{Input Lifting.} Our model processes multimodal input: the observed magnetic field and a scalar vector. We lift these inputs into high-dimensional feature spaces, denoted by \({v}_0\) and \(u_0\), respectively.

We lift \(\mathbf{B}_{\mathrm{obs}}\) via the lifting layer $P_1$, which is a Multilayer Perceptron (MLP):
\begin{equation*}
 v_0=\; P_1\bigl(\mathbf{B}_{\mathrm{obs}}\bigr),
\end{equation*}
In our application, this MLP consists of two fully connected layers. The activation function of our MLP is the Gaussian Error Linear Unit (GELU)\cite{hendrycks2023}. 
    
The scalar vector $\textbf{L}$,  is lifted by another lifting layer \(P_2\), which is also an MLP:
   \begin{equation*}
    u = P_2\bigl(\textbf{L}),
    \end{equation*}

where \(\Lvec = (\Lx,\Ly,\Lz)\) denotes the vector of actual physical lengths
in the \(x\), \(y\), and \(z\) directions, respectively. $P_2$ is the same structure as the MLP in $P_1$. The only difference is that the input for $P_2$ is a 1D vector while the input for $P_1$ is 2D images. The output $u$ is then reshaped to match the dimensions of $v_0$. 

\textbf{ECA with Dilated Convolution.} We further refine $u$ using an ECA module with a DC block:
\begin{equation*}
    u_0 = \sigma \Bigl(\mathrm{Conv1D}\bigl(\mathrm{AvgPool}(u)\bigr)\Bigr),
\end{equation*}
where \(\mathrm{AvgPool}(\cdot)\) denotes global average pooling, \(\mathrm{Conv1D}(\cdot)\) is a 1D convolution, and \(\sigma\) is the sigmoid activation. We then combine these refined scalar features with the lifted magnetic-field features by element-wise addition:
\begin{equation*}
z_0 = v_0 + u_0.
\end{equation*}

\textbf{Projection Layer.} Next, $z_0$ is passed through $n$ FNO layers, which can refer to Equations \eqref{eq:layer1}
and \eqref{eq:layer2}. As a result, we obtain the final feature vector $z_{n+1}$.  The projection layer $Q$ maps $z_{n+1}$ back to the 3D magnetic field:
\begin{equation*}
\mathbf{B} \;=\; Q\bigl(z_{n+1}\bigr),
\end{equation*}
where \(Q\) is an MLP projecting the hidden dimension back to the desired channel dimension.

\subsection{Physics-informed Loss}
To enforce our prediction and maintain physics consistency, we incorporate physics-informed loss by following definitions from \cite{jarolim2023} that constrain the field to be both divergence-free and force-free.

\textbf{Divergence-free Loss}
\begin{equation}\label{eq:physics loss 1}
\mathcal{L}_{\mathrm{div}}
= \frac{1}{N} \sum_{i=1}^{N} 
\bigl\lvert \nabla \cdot \mathbf{B}\bigr\rvert^2,
\end{equation}
where \(\mathbf{B}\) is the predicted 3D magnetic field. 

\textbf{Force-free Loss}

\begin{equation}\label{eq:physics loss 2}
\mathcal{L}_{\mathrm{ff}}
= \frac{1}{N} \sum_{i=1}^{N}
\frac{\bigl\|(\nabla \times \mathbf{B}) \times \mathbf{B}\bigr\|^2}
     {\bigl\|\mathbf{B}\bigr\|^2 + \epsilon},
\end{equation} where \(\epsilon >0\) is a small constant for numerical stability.

By combing \eqref{eq:physics loss 1} and \eqref{eq:physics loss 2}, our physics-informed loss follows:
\begin{equation}\label{eq:physics loss final}
\mathcal{L}_{\mathrm{physics}}
= \lambda_{\mathrm{div}}\,\mathcal{L}_{\mathrm{div}}
\;+\;\lambda_{\mathrm{ff}}\,\mathcal{L}_{\mathrm{ff}},
\end{equation}
where \(\lambda_{\mathrm{div}}, \lambda_{\mathrm{ff}}>0\) are the hyperparameters balancing the two terms.

We also use the H1 loss, used in the original FNO \cite{li2021}, to represent our data loss during the training. Specifically, we have: 
\vspace{-0.2cm}

\begin{equation} \label{eq:data loss}
\mathcal{L}_{\mathrm{data}} = \left\| \mathbf{B} - \mathbf{B}_{\text{true}} \right\|_{2}^2 + \sum_{i=1}^{d} \left\| \frac{\partial \mathbf{B}}{\partial x_i} - \frac{\partial \mathbf{B}_{\text{true}}}{\partial x_i} \right\|_{2}^2.
\end{equation}
Here, $d$ is the number of spatial dimensions. In our case, $d=3$ where $\partial x_1=\partial x$, $\partial x_2=\partial y$, $\partial x_3=\partial z$.   Our final loss function is defined by adding \eqref{eq:physics loss final} and \eqref{eq:data loss} together
\begin{equation}
\mathcal{L} = \mathcal{L}_{\mathrm{data}}+\mathcal{L}_{\mathrm{physics}}.
\end{equation}

\subsection{Two-phase Model}
We employ a two-phase model for training:

\textbf{Phase 1:}  Train model according to the procedure described in Figure~\ref{fig:PIANO}. We use 2D boundary data together with the physics-informed loss to train the network.

\textbf{Phase 2:} Once we obtained the model from \textbf{Phase 1}, we retrain the model using the prediction of the model from \textbf{Phase 1} as input, since the prediction is in 3D dimensions, which provides more information compared to the original input of 2D boundary conditions. 

\section{Experiment}
\subsection{Dataset} 170 active regions of the ISEE NLFFF dataset \cite{kusano2021} from 2010 to 2016, which is available from \url{https://hinode.isee.nagoya-u.ac.jp/nlfff_database/}, are used for model training and test.  143 active regions data from  2010 to 2014 are used for training, while the remaining 27 active regions data from 2015 to 2016 are used as a test set.  \begin{figure}[!htb]
    \centering
    \includegraphics[width=0.5\textwidth]{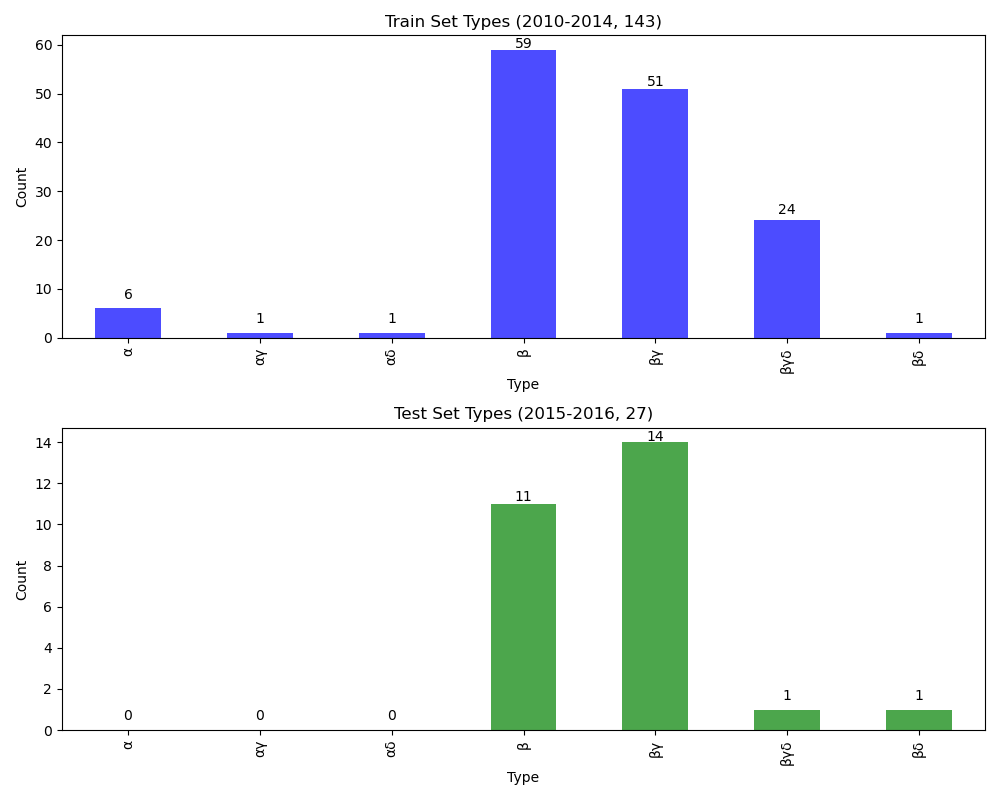}\vspace{-0.2cm}
    \caption{Active regions type distribution of the dataset. The top subfigure illustrates the counts of different types in the training set, which has 143 samples from 2010 to 2014. The bottom subfigure shows the distribution of types in the test set, which has 27 samples from 2015 to 2016. The majority of the data belongs to the \(\beta\) and \(\beta\gamma\) types, with smaller counts for others.}
    \vspace{-0.4cm}
    \label{fig:Distribution}
\end{figure} 
The dataset includes various types of active regions, with the detailed distribution provided in Fig.~\ref{fig:Distribution}. As preprocessing, $\mathbf{B}_x$, $\mathbf{B}_y$, and $\mathbf{B}_z$ are denoised by removing the magnetic field within the range of -10 to 10 Gauss and then standardized using Z-score normalization. In our experiments, we aim to learn the operator in \eqref{eq: task}, where the observed photospheric boundary measurements $\mathbf{B}^{\text{obs}} \in \mathbb{R}^{3\times 257 \times 513}$ to the extrapolated $\mathbf{B} \in \mathbb{R}^{3\times 128 \times 257 \times 513}$. 

\subsection{Experiment Setting}

\textbf{Baseline Methods.} We compare our performance with many SOTA FNO-based methods, including the FNO \cite{li2021}, GLFNO \cite{du2024global}, UFNO \cite{wen2022}, GeoFNO \cite{li2024}, GNOT \cite{hao2023}, FNOMIO \cite{JIANG2024110392}, and PINO \cite{Jeon2025}. 

\textbf{Evaluation metric.} For evaluation, we calculate the R-square ($R^2$), Relative Error (RE), Mean Squared Error (MSE), Mean Absolute Error (MAE), Peak Signal-to-Noise Ratio (PSNR), and Structural Similarity Index Measure (SSIM). 

\begin{figure*}[htbp]
    \centering
    \begin{subfigure}[t]{0.45\textwidth}
        \centering
        \includegraphics[width=\textwidth]{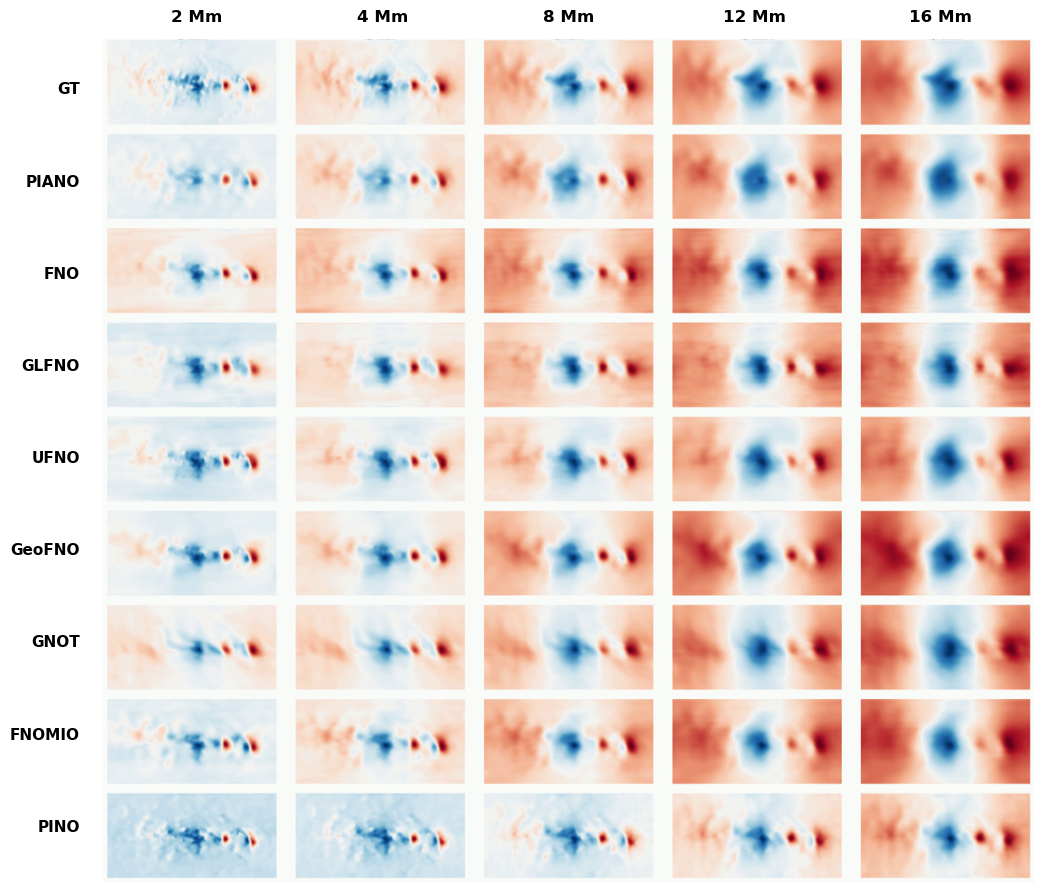}

        \caption{$\mathbf{B}_x$ Visualization}
        \label{fig:result_bx}
    \end{subfigure}
    \hfill
    \begin{subfigure}[t]{0.47\textwidth}
        \centering
        \includegraphics[width=\textwidth]{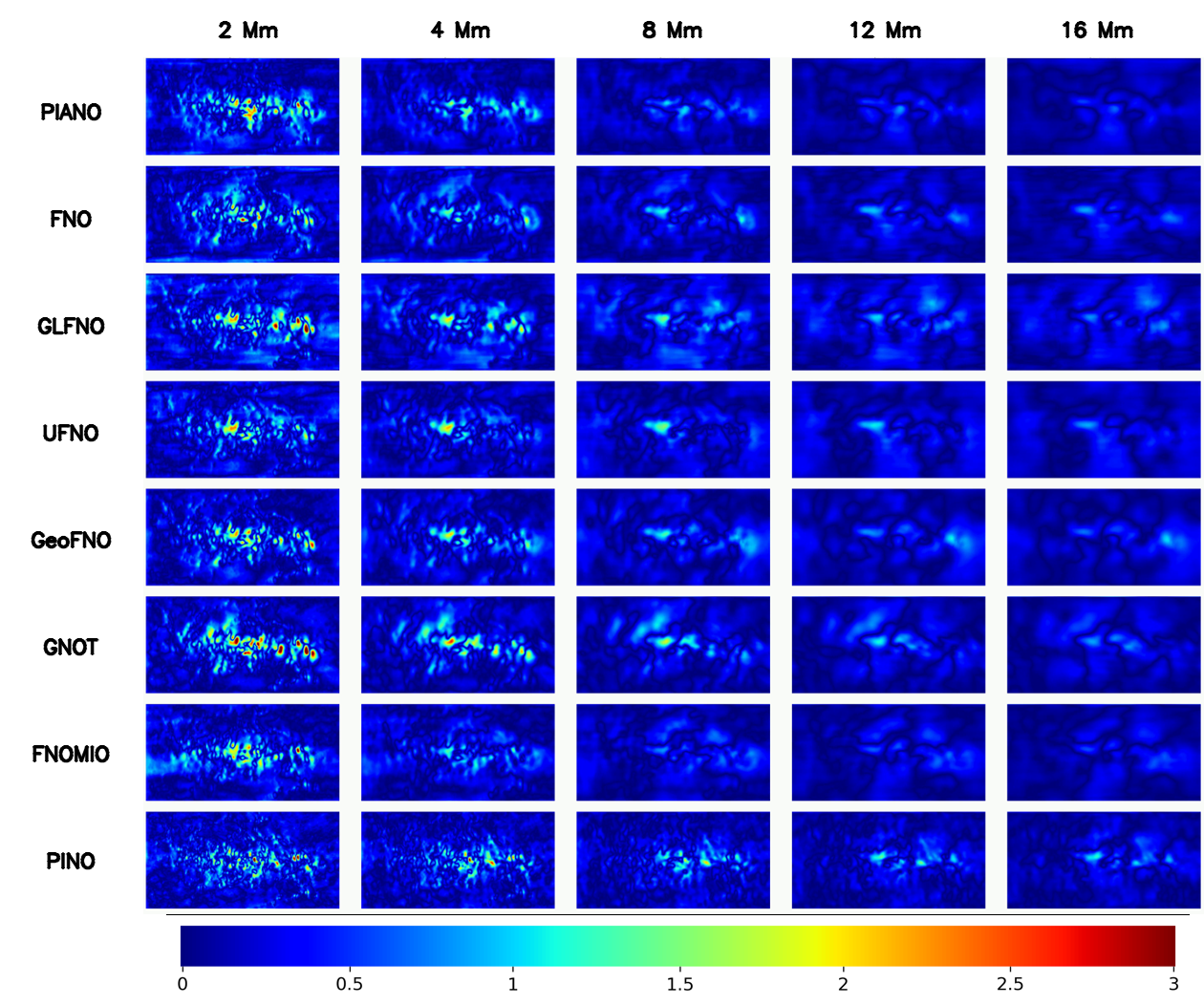}

        \caption{$\mathbf{B}_{x}$ Error Map}
        \label{fig:error_bx}
    \end{subfigure}


    \begin{subfigure}[t]{0.45\textwidth}
        \centering
        \includegraphics[width=\textwidth]{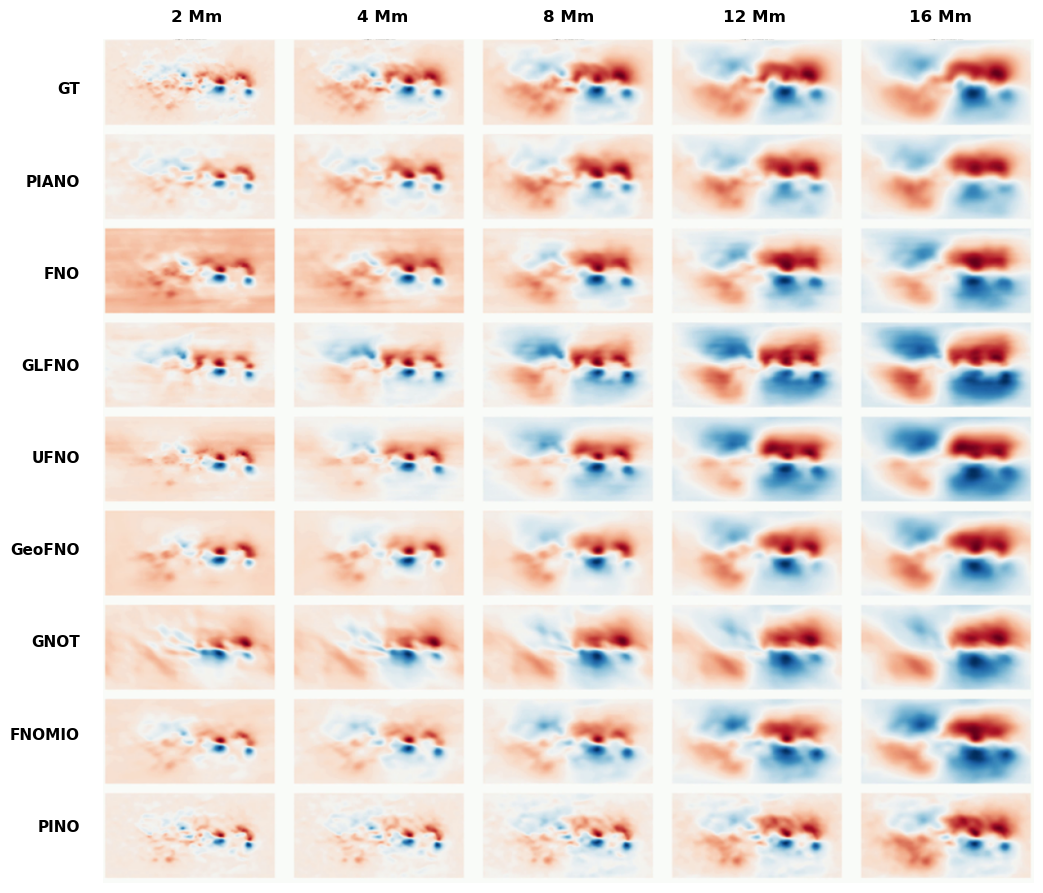}

        \caption{$\mathbf{B}_y$ Visualization}
        \label{fig:result_by}
    \end{subfigure}
    \hfill
    \begin{subfigure}[t]{0.47\textwidth}
        \centering
        \includegraphics[width=\textwidth]{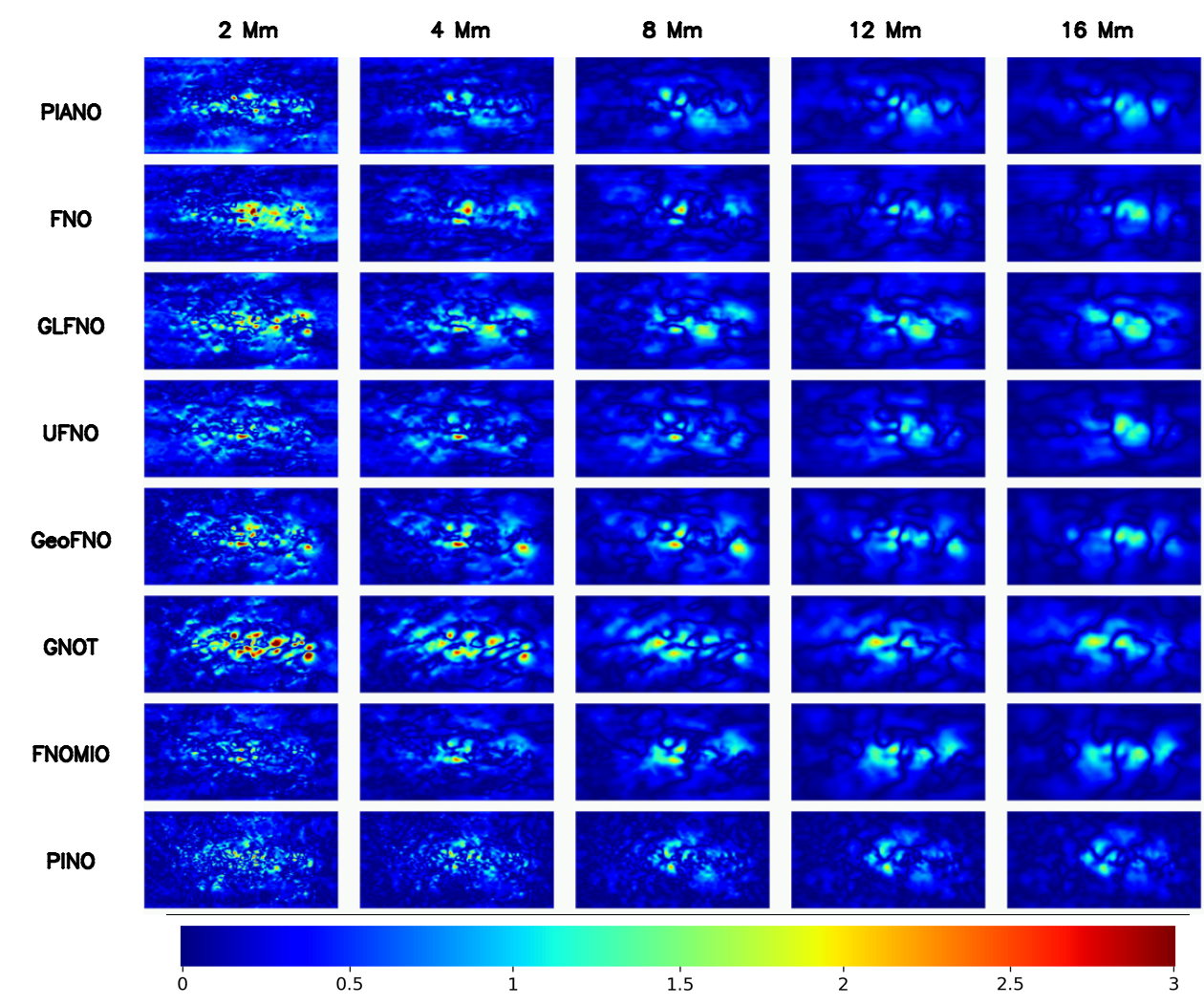}

        \caption{$\mathbf{B}_y$ Error Map}
        \label{fig:error_by}
    \end{subfigure}


    \begin{subfigure}[t]{0.45\textwidth}
        \centering
        \includegraphics[width=\textwidth]{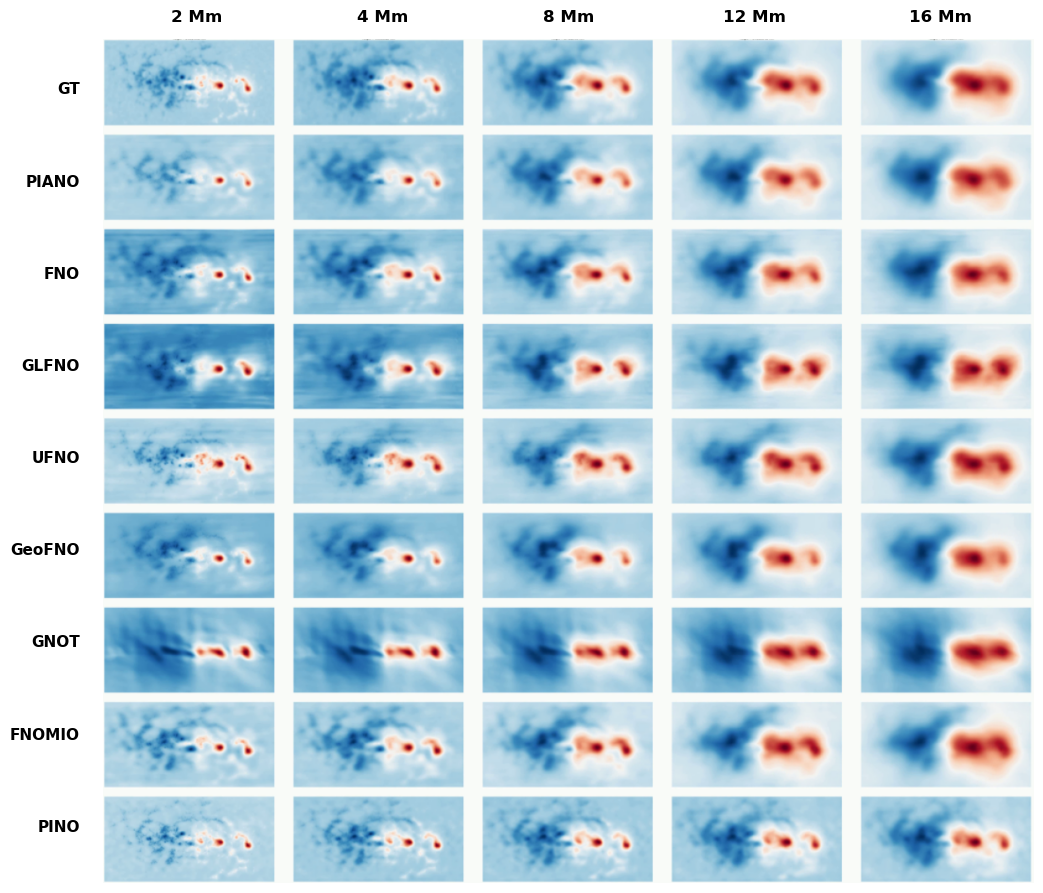}

        \caption{$\mathbf{B}_z$ Visualization}
        \label{fig:result_bz}
    \end{subfigure}
    \hfill
    \begin{subfigure}[t]{0.47\textwidth}
        \centering
        \includegraphics[width=\textwidth]{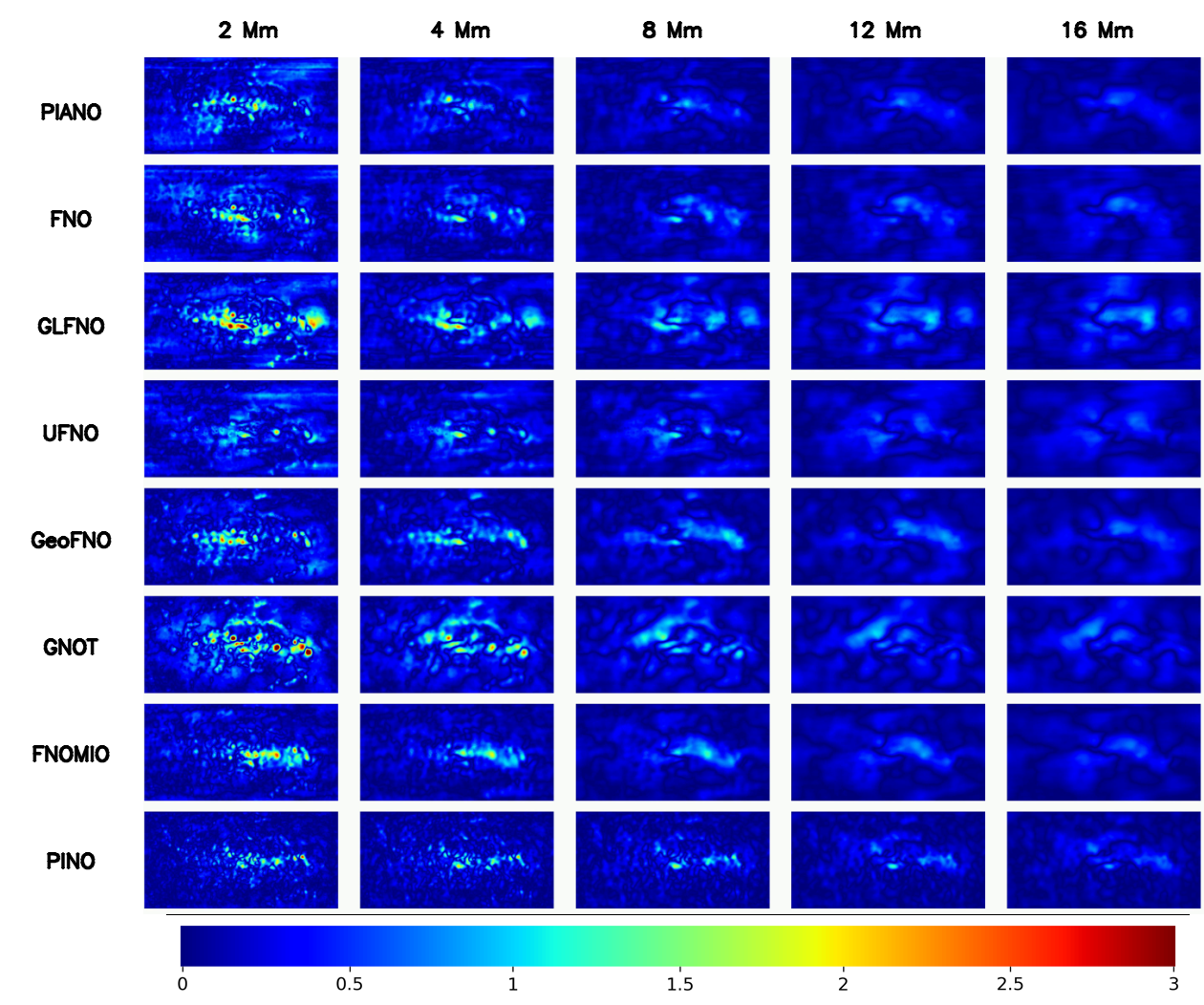}

        \caption{$\mathbf{B}_z$ Error Map}
        \label{fig:error_bz}
    \end{subfigure}
    \vspace{-0.1cm}
    \caption{Comparison of the visualization and error maps of $\mathbf{B}_x$, $\mathbf{B}_y$ and $\mathbf{B}_z$ across different neural operator models. Subfigures (a), (c), and (e) depict the visualizations of $\mathbf{B}_x$, $\mathbf{B}_y$ and $\mathbf{B}_z$, respectively, at different heights (2 Mm, 4 Mm, 8 Mm, 12 Mm, 16 Mm) for the ground truth (GT) and other neural operator models. Subfigures (b), (d), and (f) show the corresponding error maps, highlighting the prediction errors for each model compared to the ground truth.}
    \label{fig:Extrapolations_comparison}
\end{figure*}

\begin{table}[!htbp]
    \caption{Test performance comparison of PIANO, FNO, GLFNO, UFNO, GeoFNO, GNOT, FNOMIO, and PINO with different evaluation metrics.}
    \centering
    \setlength{\tabcolsep}{3pt}
    \resizebox{\columnwidth}{!}{
    \begin{tabular}{@{}cccccccc@{}}
        \toprule
        Model & Component & $R^2$ \textuparrow & RE \textdownarrow & MSE \textdownarrow & MAE \textdownarrow & PSNR \textuparrow & SSIM \textuparrow\\
        \midrule
        PIANO & $\mathbf{B}_x$ & \textbf{0.9401} & \textbf{0.2470} & \textbf{0.0599} & \textbf{0.1616} & \textbf{45.4539} & \textbf{0.9503}\\
              & $\mathbf{B}_y$ & \textbf{0.9315} & \textbf{0.2909} & \textbf{0.0685} & \textbf{0.1717} & \textbf{44.8571} & \textbf{0.9403}\\
              & $\mathbf{B}_z$ & \textbf{0.9733} & \textbf{0.2050} & \textbf{0.0267} & \textbf{0.1083} & \textbf{48.9912} & \textbf{0.9716}\\
        \midrule
        FNO   & $\mathbf{B}_x$ & 0.9270 & 0.2755 & 0.0730 & 0.1802 & 44.5864 & 0.9408\\
              & $\mathbf{B}_y$ & 0.9178 & 0.3229 & 0.0822 & 0.1904 & 44.0375 & 0.9246\\
              & $\mathbf{B}_z$ & 0.9666 & 0.2315 & 0.0334 & 0.1216 & 48.0214 & 0.9663\\
        \midrule
        GLFNO & $\mathbf{B}_x$ & 0.8690 & 0.3954 & 0.1301 & 0.2582 & 41.9649 & 0.8930\\
              & $\mathbf{B}_y$ & 0.8445 & 0.4757 & 0.1555 & 0.2283 & 41.3495 & 0.8558\\
              & $\mathbf{B}_z$ & 0.9309 & 0.3586 & 0.0691 & 0.1881 & 44.7740 & 0.9174\\
        \midrule
         UFNO  & $\mathbf{B}_x$ & 0.9350 & 0.2734 & 0.0650 & 0.1782 & 45.0748 & 0.9420\\
              & $\mathbf{B}_y$ & 0.9310 & 0.3051 & 0.0690 & 0.1799 & 44.7790 & 0.9347\\
              & $\mathbf{B}_z$ & 0.9728 & 0.2158 & 0.0272 & 0.1130 & 48.8829 & 0.9708\\
        \midrule
        GeoFNO& $\mathbf{B}_x$ & 0.9318 & 0.2632 & 0.0682 & 0.1720 & 44.8576 & 0.9437\\
              & $\mathbf{B}_y$ & 0.9269 & 0.3020 & 0.0731 & 0.1782 & 44.5406 & 0.9335\\
              & $\mathbf{B}_z$ & 0.9691 & 0.2176 & 0.0309 & 0.1145 & 48.3242 & 0.9707\\
        \midrule
        GNOT  & $\mathbf{B}_x$ & 0.8283 & 0.3866 & 0.1717 & 0.2533 & 40.9512 & 0.8928\\
              & $\mathbf{B}_y$ & 0.7878 & 0.5143 & 0.2122 & 0.3063 & 39.9928 & 0.8302\\
              & $\mathbf{B}_z$ & 0.8501 & 0.4548 & 0.1499 & 0.2429 & 41.5471 & 0.8803\\
        \midrule
        FNOMIO& $\mathbf{B}_x$ & 0.9378 & 0.2487 & 0.0622 & 0.1626 & 45.3081 & 0.9500\\
              & $\mathbf{B}_y$ & 0.9235 & 0.3135 & 0.0765 & 0.1847 & 44.3536 & 0.9299\\
              & $\mathbf{B}_z$ & 0.9708 & 0.2166 & 0.0292 & 0.1140 & 48.6723 & 0.9698\\
        \midrule
        PINO  & $\mathbf{B}_x$ & 0.9253 & 0.2773 & 0.0747 & 0.1809 & 44.4923 & 0.9387\\
              & $\mathbf{B}_y$ & 0.9184 & 0.3246 & 0.0816 & 0.1912 & 44.0795 & 0.9260\\
              & $\mathbf{B}_z$ & 0.9676 & 0.2250 & 0.0332 & 0.1182 & 48.2130 & 0.9664\\
        
        \bottomrule
    \end{tabular}}
    \vspace{-0.3cm}
    \label{tab:Performance}
\end{table}

\begin{table*}[ht]\vspace{-0.2cm}
\caption{Comprehensive ablation study across magnetic field components. The best results are in bold.} 
\centering
\renewcommand{\arraystretch}{1}
\setlength{\tabcolsep}{4pt}
\small

\begin{tabular}{@{}c c c c @{\hspace{1.5em}} c c c c c c@{}}
\toprule
Component & FNO & ECA & Physics Loss& $R^2$ $\uparrow$ & RE $\downarrow$ & MSE $\downarrow$ & MAE $\downarrow$ & PSNR $\uparrow$ & SSIM $\uparrow$ \\
& + Scalars & with DC & with 2-Phase Training &  &  &  &  &  &  \\
\midrule

\multirow{3}{*}{$\mathbf{B}_x$}  
 & \checkmark & \xmark & \xmark & 0.9338 & 0.2605 & 0.0661 & 0.1701 & 44.9852 & 0.9452 \\
 & \checkmark & \checkmark & \xmark & 0.9371 & 0.2479 & 0.0629 & 0.1623 & 45.2552 & 0.9495 \\
 & \checkmark & \checkmark & \checkmark & \textbf{0.9401} & \textbf{0.2470} & \textbf{0.0599} & \textbf{0.1616} & \textbf{45.4539} & \textbf{0.9503} \\
\cmidrule(lr){1-10}

\multirow{3}{*}{$\mathbf{B}_y$}  
 & \checkmark & \xmark & \xmark & 0.9231 & 0.3115 & 0.0768 & 0.1835 & 44.3581 & 0.9324 \\
 & \checkmark & \checkmark & \xmark & 0.9241 & 0.3053 & 0.0759 & 0.1803 & 44.4134 & 0.9339 \\
 & \checkmark & \checkmark & \checkmark & \textbf{0.9315} & \textbf{0.2909} & \textbf{0.0685} & \textbf{0.1717} & \textbf{44.8571} & \textbf{0.9403} \\
\cmidrule(lr){1-10} 

\multirow{3}{*}{$\mathbf{B}_z$}  
 & \checkmark & \xmark & \xmark & 0.9696 & 0.2199 & 0.0304 & 0.1160 & 48.3758 & 0.9681 \\
 & \checkmark & \checkmark & \xmark & 0.9701 & 0.2166 & 0.0299 & 0.1142 & 48.6168 & 0.9678 \\
 & \checkmark & \checkmark & \checkmark & \textbf{0.9733} & \textbf{0.2050} & \textbf{0.0267} & \textbf{0.1083} & \textbf{48.9912} & \textbf{0.9716} \\
\bottomrule
\end{tabular}\vspace{-0.3cm}
\label{tab:AblationCombined}
\end{table*}

\subsection{Experimental Results}

The experimental results evaluate the performance of various neural operator models in extrapolating the magnetic field components $\mathbf{B}_x$, $\mathbf{B}_y$ and $\mathbf{B}_z$ at different heights (2 Mm, 4 Mm, 8 Mm, 12 Mm, 16 Mm). Both visual comparisons and quantitative metrics highlight the strengths of our PIANO model.

\textbf{Visual Comparisons.} The visualization sample is active region NOAA 12443, which is an \(\beta\delta\) active region on 11/04, 2015. Fig.~\ref{fig:Extrapolations_comparison} provides a detailed comparison of the predicted magnetic field components and their corresponding error maps. Error maps reveal that higher errors are concentrated in low-height regions, particularly at 2 Mm and 4 Mm, with a gradual reduction as the height increases. We can observe that most of the SOTA models we selected were able to achieve good extrapolation results in the $\mathbf{B}_x$ and $\mathbf{B}_z$. However, the structure in the $\mathbf{B}_y$ is more complex compared to $\mathbf{B}_x$ and $\mathbf{B}_z$, leading to certain extrapolation errors at all the selected heights. Among the models, PIANO, GNOT, PINO, and FNOMIO exhibit better accuracy in the lower height, and PIANO shows superior accuracy, especially on $\mathbf{B}_y$, in the higher height.

\textbf{Quantitative Metrics.} We present our experimental results in Table \ref{tab:Performance}. The metrics are calculated over the three models' prediction components $\mathbf{B}_x$, $\mathbf{B}_y$, and $\mathbf{B}_z$ along three different directions. The proposed PIANO has the highest $R^2$, PSNR, and SSIM, and has the lowest RE, MSE, and MAE on all the components. While other methods also perform well on $\mathbf{B}_x$ and $\mathbf{B}_z$, they are still slightly worse than PIANO. As for the metrics of $\mathbf{B}_y$, PIANO shows a significant improvement compared to other methods. So we can conclude that PIANO outperforms other models in all metrics, indicating its great ability to generalize across different magnetic field components and heights. 

\begin{figure}[!htb]\vspace{-0.1cm}
    \centering
    \subfloat[$\beta$'s divergence-free loss]{
         \includegraphics[width=0.45\linewidth]{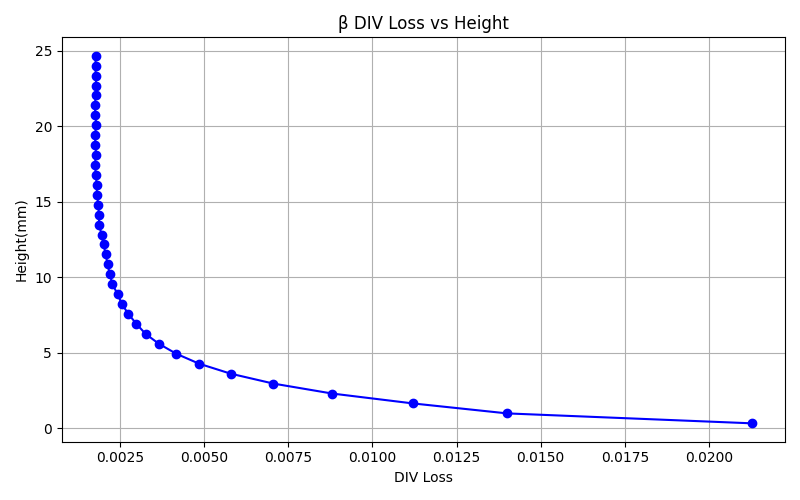}
         \label{fig:sub1}
     }
     \hfill
     \subfloat[$\beta$'s force-free loss]{
         \includegraphics[width=0.45\linewidth]{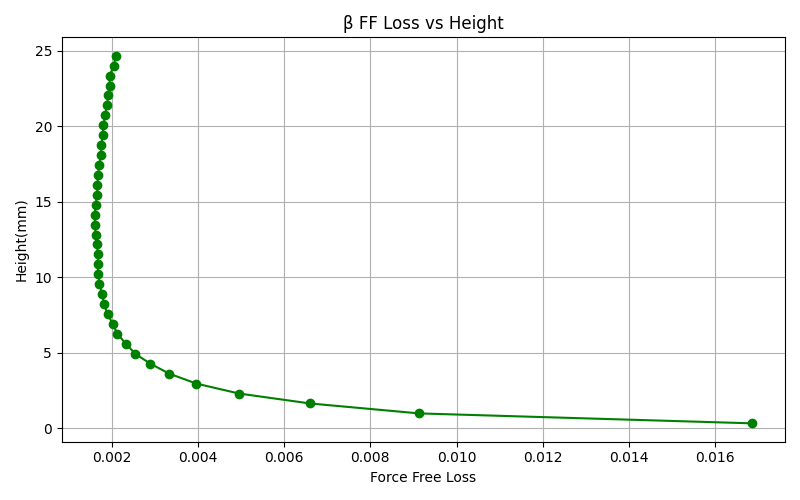}
         \label{fig:sub2}
     }


    \subfloat[$\beta\delta$'s divergence-free loss]{
        \includegraphics[width=0.45\linewidth]{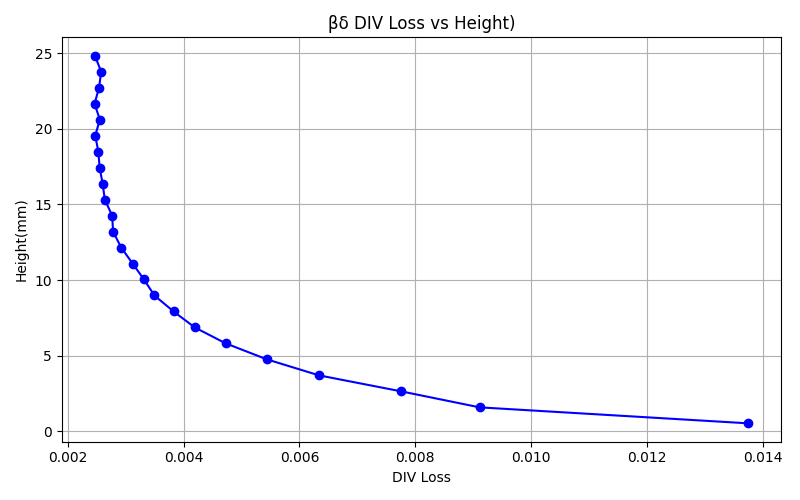}
        \label{fig:sub3}
    }
    \hfill
    \subfloat[$\beta\delta$'s force-free loss]{
        \includegraphics[width=0.45\linewidth]{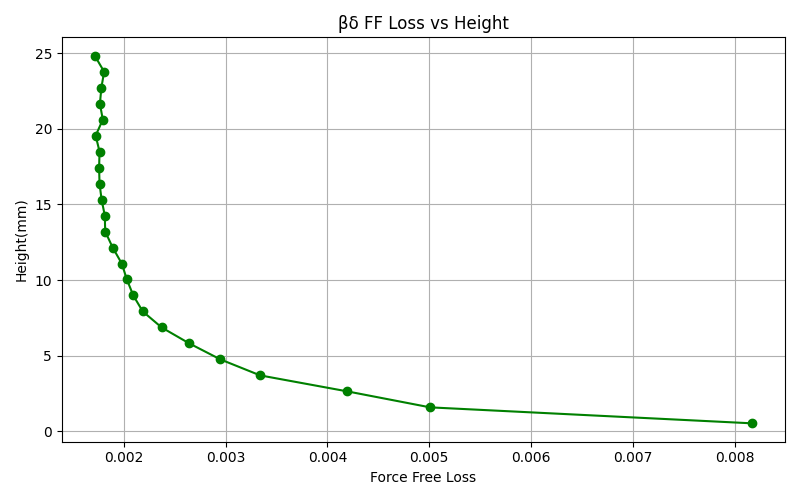}
        \label{fig:sub4}
    }


    \subfloat[$\beta\gamma$'s divergence-free loss]{
        \includegraphics[width=0.45\linewidth]{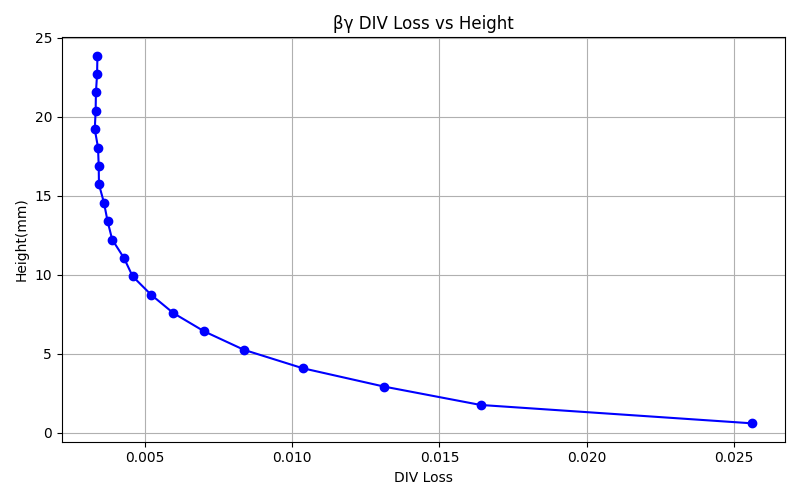}
        \label{fig:sub5}
    }
    \hfill
    \subfloat[$\beta\gamma$'s force-free loss]{
        \includegraphics[width=0.45\linewidth]{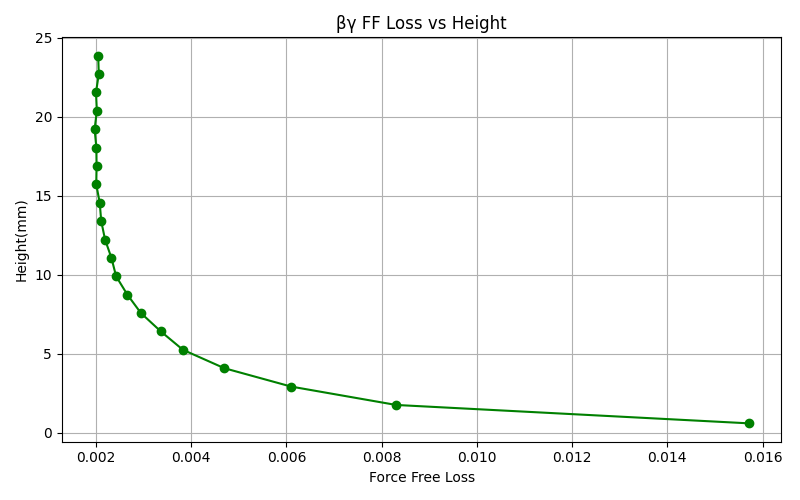}
        \label{fig:sub6}
    }


    \subfloat[$\beta\gamma\delta$'s divergence-free loss]{
        \includegraphics[width=0.45\linewidth]{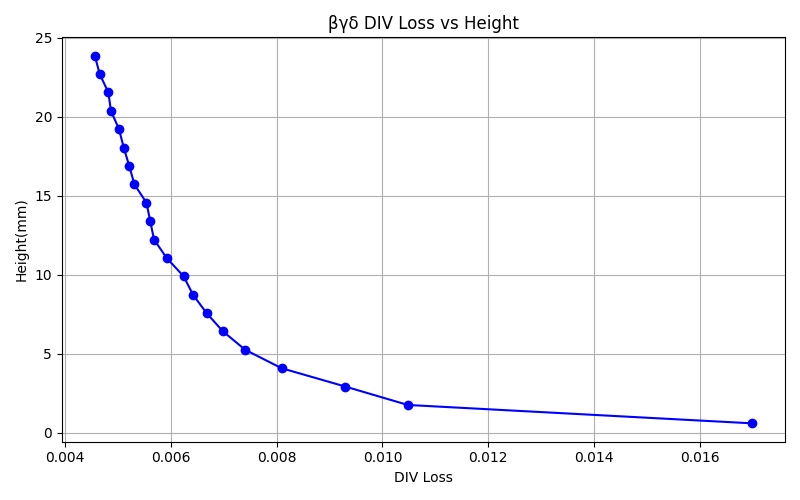}
        \label{fig:sub7}
    }
    \hfill
    \subfloat[$\beta\gamma\delta$'s force-free loss]{
        \includegraphics[width=0.45\linewidth]{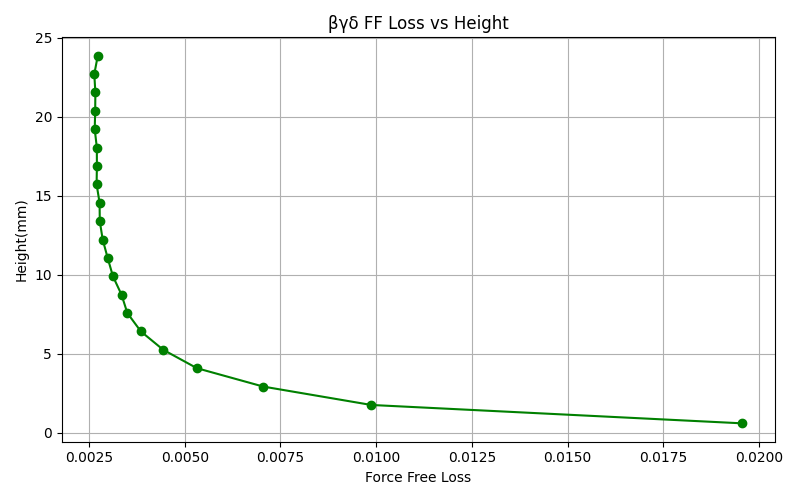}
        \label{fig:sub8}
    }

    \caption{Variation of divergence-free loss and force-free loss with height for different types of solar active regions. Subfigures (a), (c), (e) and (g) show the divergence-free loss as a function of height for \(\beta\), \(\beta\delta\), \(\beta\gamma\), and \(\beta\gamma\delta\), respectively. Subfigures (b), (d), (f) and (h) depict the corresponding force-free loss for the same types. Both losses decrease with height across all configurations, with higher losses observed at lower heights, indicating greater complexity in the magnetic field structures at lower heights.} 
    \vspace{-0.7cm}
    \label{fig:physicalloss}
\end{figure} 

\subsection{Ablation Study}

We conducted a comprehensive ablation study to verify whether our proposed method effectively improves the model's performance. The results of the ablation study are provided in Table \ref{tab:AblationCombined}. 

\textbf{Individual Component Contributions.} When \textbf{ECA with DC} block is disabled (i.e., only \textbf{FNO+Scalars} is enabled), the model shows the worst performance, particularly  $R^2$= 0.9231, RE = 0.3115, and MSE = 0.0768. This indicates that the \textbf{ECA with DC} block is critical for capturing key scalar features, enhancing the model's representational capacity. Disabling only the \textbf{Physics loss with 2-phase Training} results in better performance compared to the above case ($R^2$ improves from 0.9231 to 0.9241, MSE decreases from 0.0768 to 0.0759). But it is still suboptimal compared to the complete configuration. This demonstrates the importance of incorporating physics-informed loss for further refining the model's predictive accuracy and generalization.

\begin{figure*}[!htb]   \vspace{-0.3cm}
    \centering
    \begin{subfigure}{0.32\textwidth}
        \centering
        \includegraphics[width=\linewidth]{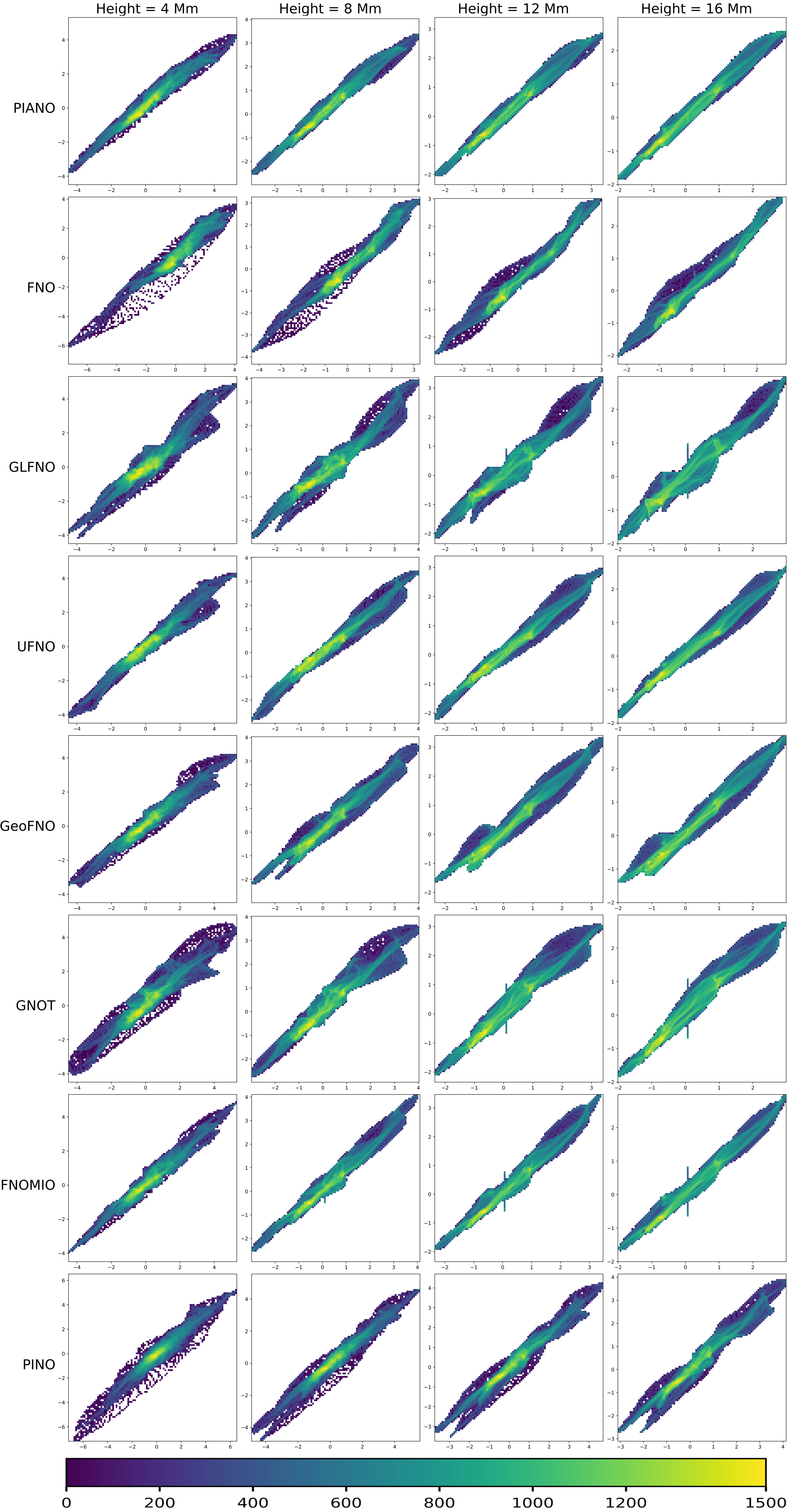}
        \caption{$\mathbf{B}_x$ 2D histogram}
    \end{subfigure}
    \begin{subfigure}{0.32\textwidth}
        \centering
        \includegraphics[width=\linewidth]{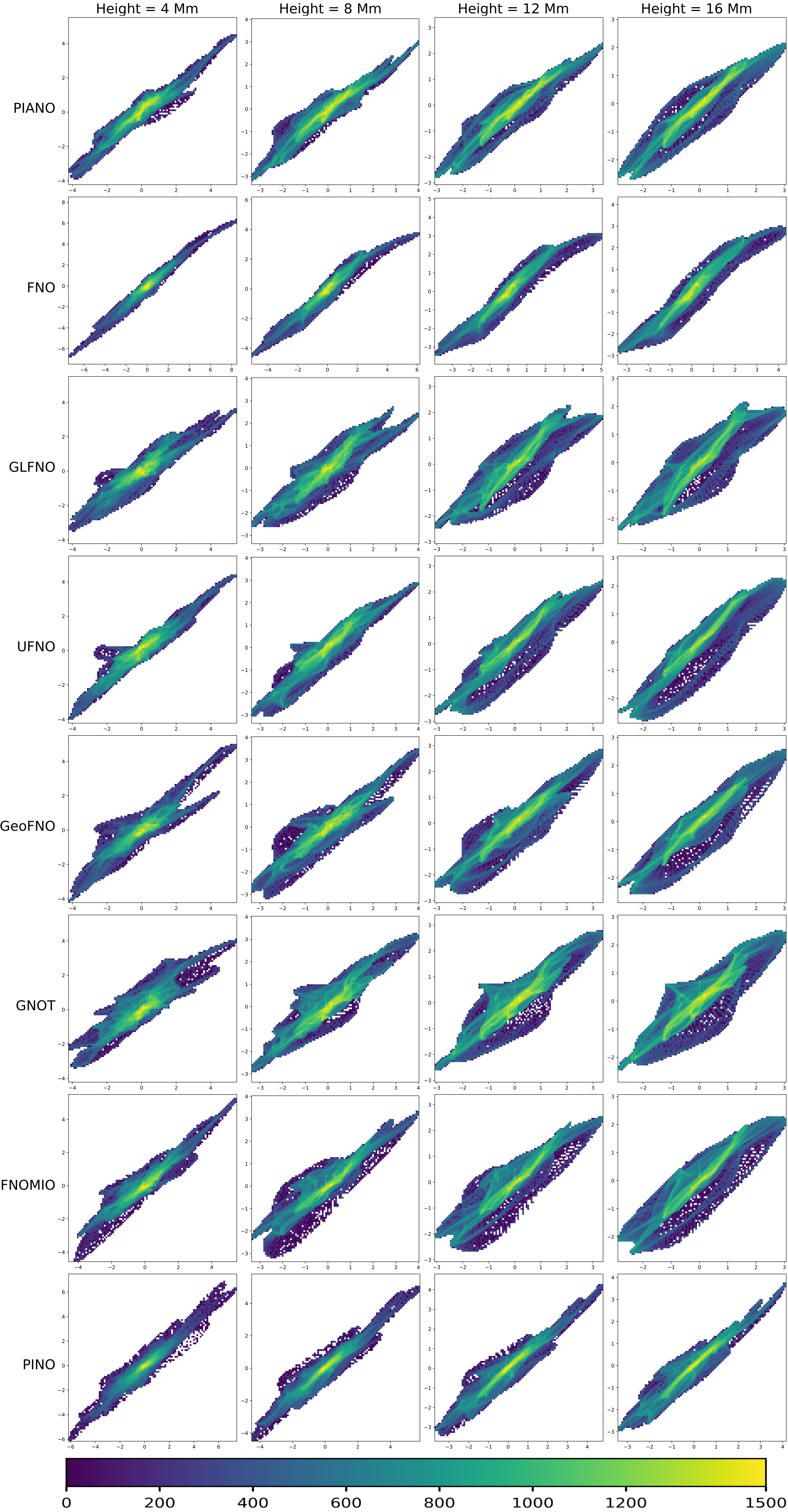}
        \caption{$\mathbf{B}_y$ 2D histogram}
    \end{subfigure}
    \begin{subfigure}{0.32\textwidth}
        \centering
        \includegraphics[width=\linewidth]{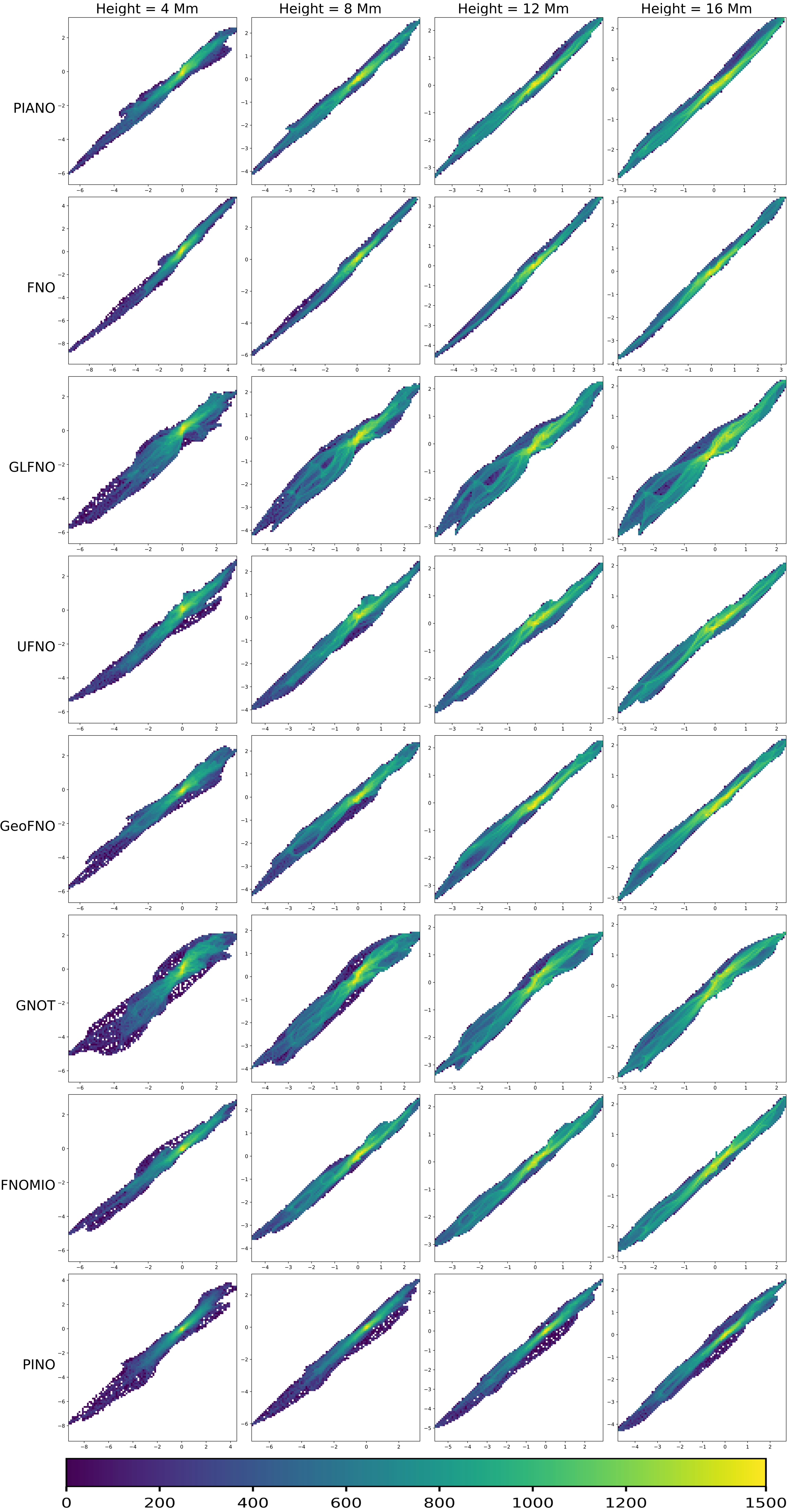}
        \caption{$\mathbf{B}_z$ 2D histogram}
    \end{subfigure}
    \vspace{-0.2cm}
    \caption{2D histograms comparing the predicted $\mathbf{B}_x$, $\mathbf{B}_y$ and $\mathbf{B}_z$ to the ground truth. at different heights (4 Mm, 8 Mm, 12 Mm, and 16 Mm) for various neural operator models. Subfigure (a) displays the 2D histograms for $\mathbf{B}_x$, subfigure (b) for $\mathbf{B}_y$, and subfigure (c) for $\mathbf{B}_z$. Each row corresponds to a model, and each column represents a specific height. The color intensity reflects the density of data points, with brighter colors indicating higher density.} \vspace{-0.4cm}
    \label{fig:2Dhistogram}
\end{figure*}

\begin{figure}[t]
    \centering
    \begin{subfigure}[b]{\linewidth}
        \centering
        \includegraphics[width=\linewidth]{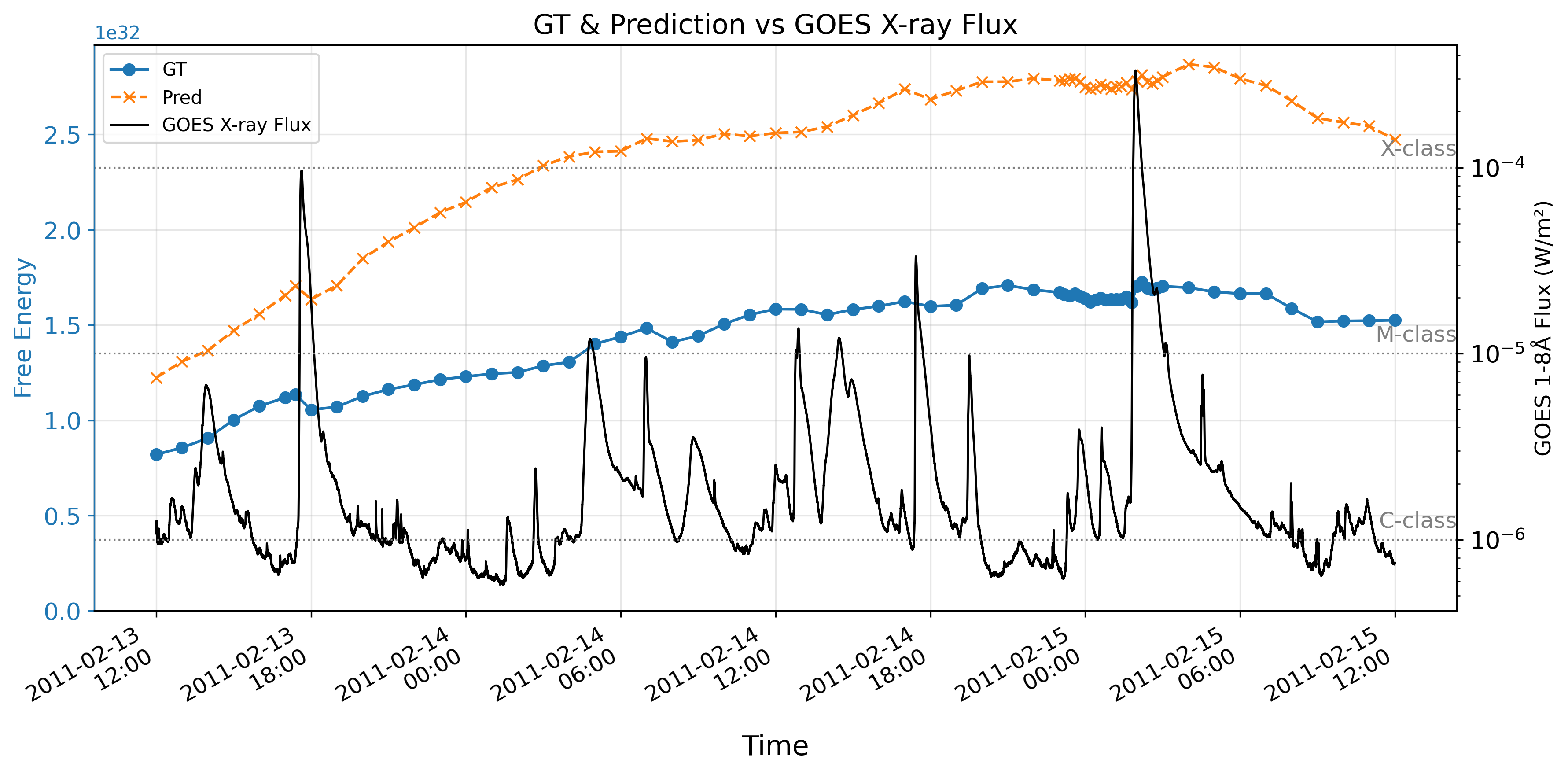}
        \vspace{-0.6cm}
        \caption{Free magnetic energy vs. GOES 1--8\,\AA\ X-ray flux.}
    \end{subfigure}
    \par\medskip
    \begin{subfigure}[b]{\linewidth}
        \centering
        \includegraphics[width=\linewidth]{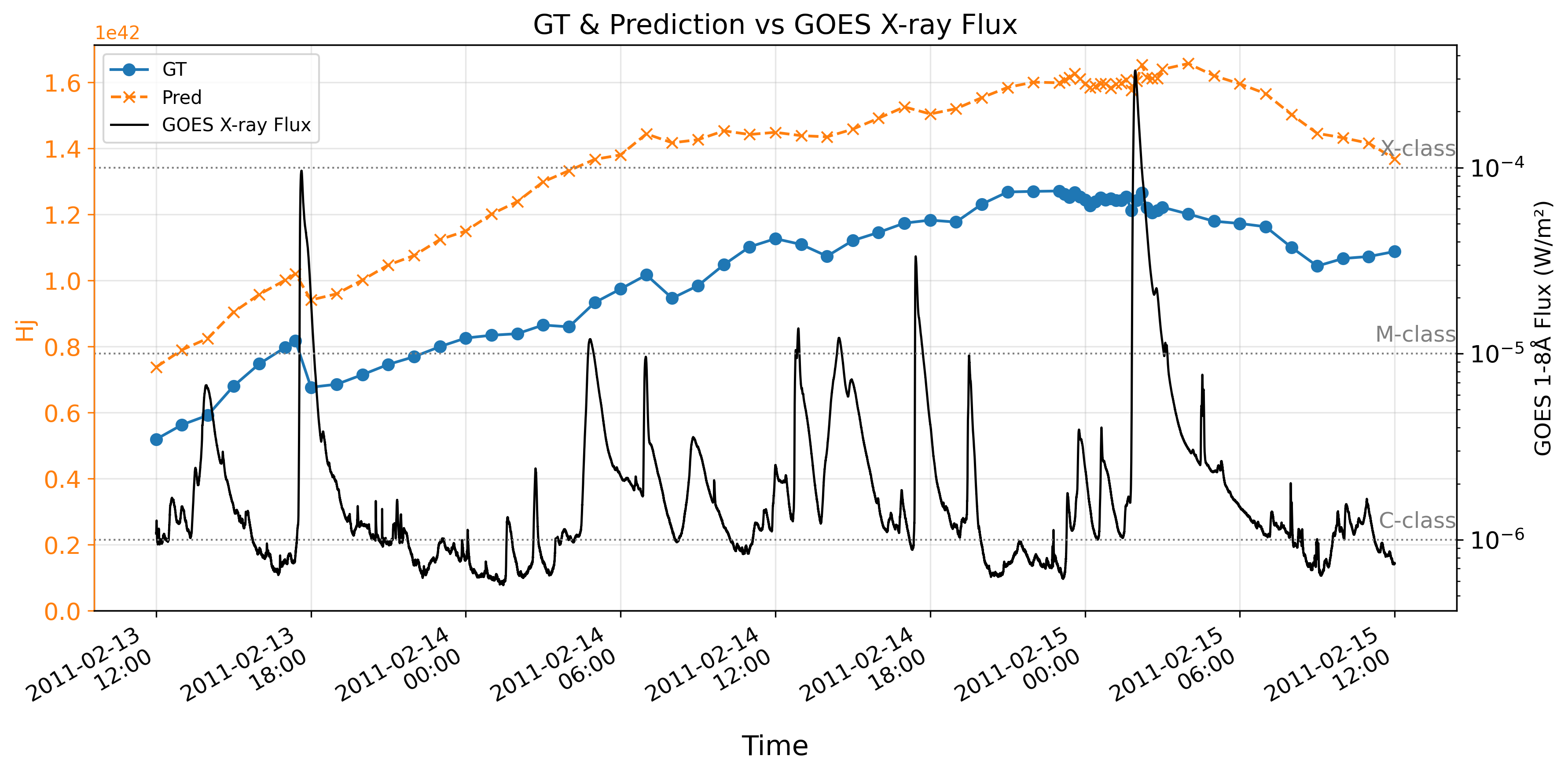}
        \vspace{-0.6cm}
        \caption{Relative magnetic helicity $H_j$ vs. GOES 1--8\,\AA\ X-ray flux.}
    \end{subfigure}
    \vspace{-0.5cm}
    \caption{Temporal evolution of free magnetic energy and relative magnetic helicity compared with the GOES 1--8\,\AA\ X-ray flux. The left axis corresponds to the magnetic quantities (GT in blue dots, PIANO in orange crosses), while the right logarithmic axis shows the GOES 1--8\,\AA\ X-ray flux with C/M/X-class thresholds (dotted gray lines). PIANO reproduces the overall growth trends, albeit with a systematic amplitude overestimation, and captures the build-up phases preceding major flares.}
    \vspace{-0.7cm}
    \label{fig:energy_helicity_goes}
\end{figure}

\textbf{Overall Performance Improvements.} The full configuration significantly improves performance across all metrics and shows substantial improvements in both error reduction (e.g., MSE drops to 0.0685 and MAE to 0.1717) and perceptual quality metrics (e.g., PSNR increases to 44.8571 and SSIM to 0.9403). This highlights the synergistic effect of combining all components, particularly the physics-informed training strategy, which not only reduces prediction errors but also preserves the spatial and structural characteristics of the data. Metrics such as PSNR and SSIM confirm that the model effectively reconstructs finer details, which are crucial for high-fidelity outputs, particularly in applications where structural integrity matters.

\subsection{Physics Evaluation}

\textbf{Physics losses.} In addition to calculating these AI/ML evaluation metrics, we also evaluate the divergence loss and force-free loss for different types of solar active regions as shown in Fig.~\ref{fig:physicalloss}. The trends of these loss curves align with \cite{jarolim2024}, showing higher values at the lower height, followed by a rapid decline beyond a certain height. The force-free condition ensures that the model is consistent with the physical assumption that the Lorentz force is negligible.  The divergence-free condition ensures the model adheres to Maxwell's equations. At lower heights (photosphere), plasma pressure and Lorentz forces are significant, leading to higher force-free loss, whereas at higher heights (corona), the low plasma $\beta$ regime, where magnetic forces dominate over thermal pressure forces, better satisfies force-free conditions. Divergence loss decreases with height, indicating a weak field gradient and smoother field structures. Generally, the balance between divergence and force loss terms matters to achieve a physically consistent solution, which was also demonstrated by our results. The traditional model tends to exhibit a gradual increase in loss with height while maintaining low values. This occurs because the model receives less information from the boundary as it moves higher. PIANO helps mitigate this degradation, reducing the impact on extrapolation at higher heights.

\textbf{2D component-wise histograms.} Fig.~\ref{fig:2Dhistogram} presents the 2D histograms comparing the $\mathbf{B}_x$, $\mathbf{B}_y$, and $\mathbf{B}_z$ to the ground truth at various heights (4 Mm, 8 Mm, 12 Mm, and 16 Mm) for different comparison models. Across all models, the $\mathbf{B}_z$ aligns closely with the ground truth, and PIANO shows the tightest clustering along the diagonal, especially at 12 Mm and 16 Mm. The alignment of predicted $\mathbf{B}_y$ with the ground truth is less precise compared to $\mathbf{B}_x$ and $\mathbf{B}_z$, with broader distributions around the diagonal. PIANO still performs better than other models, maintaining a relatively high density along the diagonal, even at 4 Mm. While the other models, such as GLFNO, GNOT, and GeoFNO, show noticeable scatter at all heights, highlighting their difficulties in capturing the complexities of $\mathbf{B}_y$. Predictions for $\mathbf{B}_z$ demonstrate better alignment with the ground truth compared to $\mathbf{B}_y$, though still slightly less accurate than $\mathbf{B}_x$. PIANO again shows the highest density of points along the diagonal, particularly at 12 Mm and 16 Mm, reflecting its robustness in extrapolating $\mathbf{B}_z$. Models like GNOT and GLFNO exhibit wider scatter at lower heights, similar to their behavior in $\mathbf{B}_x$ and $\mathbf{B}_y$. PINO shows a good alignment at lower heights. But at 12 and 16 Mm, it is less accurate than the PIANO, especially in $\mathbf{B}_x$. For $\mathbf{B}_x$ and $\mathbf{B}_z$, all models show better alignment with the ground truth as the height increases, indicating that predictions are more challenging at lower heights due to the increased complexity of magnetic field structures. However, predicting $\mathbf{B}_y$ at higher heights is more challenging than at lower heights, possibly because the magnetic field structures of $\mathbf{B}_y$ at higher heights are more complex compared to the other two directions, making the prediction more difficult. PIANO consistently outperforms other models across all directions and heights, as evidenced by its tighter clustering along the diagonal. This is particularly evident for $\mathbf{B}_y$. Overall, $\mathbf{B}_y$ predictions are the most challenging for all models, exhibiting the greatest deviations from the diagonal. This aligns with the quantitative metrics, which also indicate higher errors for $\mathbf{B}_y$.

\textbf{Free magnetic energy and relative magnetic helicity vs. GOES X-ray flux.} Fig.~\ref{fig:energy_helicity_goes} shows the temporal evolution of the free magnetic energy and magnetic helicity in NOAA AR~11158. As shown in Fig.~\ref{fig:energy_helicity_goes}a, the temporal evolution of the free magnetic energy $E_{\mathrm{free}}$ predicted by PIANO closely follows GT: both exhibit a sustained build-up phase before the major GOES 1–8\,\AA\ X-ray peaks, followed by a plateau or slight decline around the flare maxima. PIANO shows a systematic amplitude overestimation but preserves the overall smooth, physically plausible trend without spurious oscillations. Correlation and lag analyses between $E_{\mathrm{free}}$ and the GOES flux indicate significant positive correlations and comparable lead–lag behavior for PIANO and GT, suggesting that the model captures the pre-flare energy accumulation phase and thus retains potential early-warning value. The relative magnetic helicity $H_j$ (Fig.~\ref{fig:energy_helicity_goes}b) exhibits an almost monotonic accumulation throughout the time window and does show sharp drops at the flare peaks. PIANO reproduces the global trend with a modest upward bias in magnitude.

\section{Conclusion}
We introduced PIANO, a \textbf{P}hysics-\textbf{i}nformed~\textbf{A}ttention-enhanced~Fourier~\textbf{N}eural~\textbf{O}perator, for efficient and accurate NLFFF extrapolation by integrating multimodal input with attention and incorporating physics-informed loss. We evaluate the performance of our model on the ISEE NLFFF dataset by comparing state-of-the-art neural operators. From an AI/ML perspective, we have the highest accuracy among all baseline models. We also evaluate the performance of our model through a physics perspective, where our PIANO achieves the best physics consistency across various solar active regions.

\section{Acknowledgment}
We thank the NASA SDO team for their support with the HMI data. ISEE Database for Nonlinear Force-Free Field of Solar Active Regions $(doi: 10.34515/DATA\_HSC-00000)$ was developed by the Hinode Science Center, Institute for Space-Earth Environmental Research (ISEE), Nagoya University. This work was supported by NASA grant 80NSSC24M0174.

\bibliographystyle{IEEEtran} 
\bibliography{refs} 

\newpage

\onecolumn
\appendix
\section*{FNO Schematic Diagram}
\label{appendix:FNO}

\begin{figure}[htbp]
     \centering
    \includegraphics[width=\textwidth]{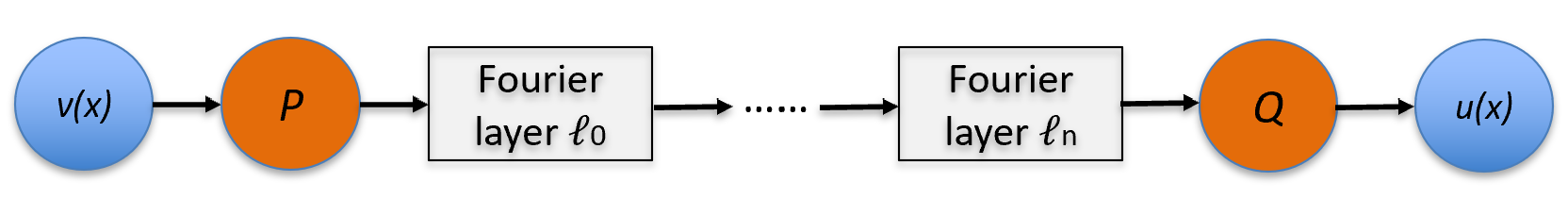}
     \caption{Architecture of FNO.  $v(x)$ is the input function and is lifted to a higher dimension channel space by a fully connected neural network $P$. $u(x)$ is the output function, which is projected back to the target dimension by a fully connected neural network $Q$. The Fourier layer is the same as we showed in Fig.~\ref{fig:PIANO}.}
     \label{fig:FNO}
    \vspace{0.2em} 
    \centering
    
 \end{figure}

 \section*{Metrics Calculation}
 \label{appendix:metrics}

 \subsection{$R^2$ (Coefficient of Determination)}
 The $R^2$ metric measures the proportion of variance in the ground truth that is predictable from the model. It is defined as:
 \[
 R^2 = 1 - \frac{\sum_{i=1}^n (y_i - \hat{y}_i)^2}{\sum_{i=1}^n (y_i - \bar{y})^2},
 \]
 where $\bar{y} = \frac{1}{n} \sum_{i=1}^n y_i$ is the mean of the ground truth values.

 \subsection{Mean Squared Error (MSE)}
 The MSE measures the average squared difference between the predicted and true values:
 \[
\text{MSE} = \frac{1}{n} \sum_{i=1}^n (y_i - \hat{y}_i)^2.
 \]

 \subsection{Mean Absolute Error (MAE)}
 The MAE is the average of the absolute differences between the predicted and true values:
 \[
 \text{MAE} = \frac{1}{n} \sum_{i=1}^n |y_i - \hat{y}_i|.
 \]

 \subsection{Relative Error}
 The Relative Error metric calculates the ratio of the prediction error to the true value, averaged over all samples:
 \[
 \text{Relative Error} = \frac{1}{n} \sum_{i=1}^n \frac{|y_i - \hat{y}_i|}{|y_i|}.
 \]

\subsection{Peak Signal-to-Noise Ratio (PSNR)}
The PSNR is a logarithmic metric that evaluates the quality of a signal relative to the maximum possible value of the signal. For data normalized between 0 and 1, it is defined as:
\[
\text{PSNR} = 10 \cdot \log_{10} \left(\frac{\text{MAX}^2}{\text{MSE}}\right),
\]
where $\text{MAX}$ is the maximum possible value of $y_i$ (typically 1 for normalized data).

\subsection{Structural Similarity Index Measure (SSIM)}
The SSIM evaluates the similarity between two images, considering luminance, contrast, and structure. It is defined as:
\[
\text{SSIM}(x, y) = \frac{(2 \mu_x \mu_y + C_1)(2 \sigma_{xy} + C_2)}{(\mu_x^2 + \mu_y^2 + C_1)(\sigma_x^2 + \sigma_y^2 + C_2)},
\]
where $\mu_x, \mu_y$ are the means of $x$ and $y$, $\sigma_x^2, \sigma_y^2$ are the variances of $x$ and $y$, $\sigma_{xy}$ is the covariance between $x$ and $y$, $C_1, C_2$ are small constants to stabilize the division.

\end{document}